\documentclass{article}

\usepackage{microtype}
\usepackage{graphicx}
\usepackage{subfigure}
\usepackage{booktabs} 

\usepackage{hyperref}

\usepackage[accepted]{icml2024}

\usepackage{amsmath}
\usepackage{amssymb}
\usepackage{mathtools}
\usepackage{amsthm}

\usepackage[capitalize,noabbrev]{cleveref}

\theoremstyle{plain}

\theoremstyle{definition}

\theoremstyle{remark}

\usepackage[textsize=tiny]{todonotes}

\usepackage{bm} 
\usepackage{bbm} 
\usepackage{comment}

\newcommand{\R}{\mathbb{R}}

\newcommand{\sood}{s_{\text{ood}}}
\newcommand{\scp}{s_{\text{cp}}}

\DeclarePairedDelimiterX{\norm}[1]{\lVert}{\rVert}{#1}
\DeclarePairedDelimiterX{\abs}[1]{\lvert}{\rvert}{#1}

\definecolor{beau-vert}{RGB}{0,102,51}

\everypar{\looseness=-1}

\icmltitlerunning{Conformal OOD}

\def\secref#1{section~\ref{#1}}

\def\eqref#1{equation~(\ref{#1})}

\def\1{\bm{1}}

\def\didcal{\mathcal{D}_{id}^{cal}}
\def\didtrain{\mathcal{D}_{id}^{train}}
\def\didtest{\mathcal{D}_{id}^{test}}
\def\dood{\mathcal{D}_{ood}}

\def\fpr{\text{FPR}}

\def\umarg{\widehat{u}^\text{marg}}
\def\ucc{\widehat{u}^\text{cc}}

\def\vh{{\bm{h}}}

\def\vx{{\bm{x}}}

\def\mH{{\bm{H}}}

\DeclareMathAlphabet{\mathsfit}{\encodingdefault}{\sfdefault}{m}{sl}
\SetMathAlphabet{\mathsfit}{bold}{\encodingdefault}{\sfdefault}{bx}{n}

\def\sX{{\mathcal{X}}}

\def\prob#1{\mathcal{P}_{#1}}

\begin{document}

\twocolumn[
\icmltitle{Out-of-Distribution Detection Should Use Conformal Prediction \\ (and Vice-versa?)}



\icmlsetsymbol{equal}{*}

\begin{icmlauthorlist}
\icmlauthor{Paul Novello}{equal,irt,aniti}
\icmlauthor{Joseba Dalmau}{equal,irt,aniti}
\icmlauthor{Leo Andeol}{sncf,imt}
\end{icmlauthorlist}

\icmlaffiliation{irt}{IRT Saint Exupery, France}
\icmlaffiliation{aniti}{Artificial and Natural Intelligence Toulouse Institute, France}
\icmlaffiliation{sncf}{SNCF, France}
\icmlaffiliation{imt}{Mathematical Institute of Toulouse, France}

\icmlcorrespondingauthor{Paul Novello}{paul.novello@irt-saintexupery.com}

\icmlkeywords{Machine Learning, ICML}

\vskip 0.3in
]



\printAffiliationsAndNotice{\icmlEqualContribution} 

\begin{abstract}
    Research on Out-Of-Distribution (OOD) detection focuses mainly on building scores that efficiently distinguish OOD data from In Distribution (ID) data. 
    On the other hand, Conformal Prediction (CP) uses non-conformity scores to construct prediction sets with probabilistic coverage guarantees. 
    In this work, we propose to use CP to better assess the efficiency of OOD scores. 
    Specifically, we emphasize that in standard OOD benchmark settings, evaluation metrics can be overly optimistic due to the finite sample size of the test dataset.
    Based on the work of \cite{bates_testing_2022}, we define new \emph{conformal AUROC} and \emph{conformal FRP@TPR95} metrics, 
    which are corrections that provide probabilistic conservativeness guarantees on the variability of these metrics.
    We show the effect of these corrections on two reference OOD and anomaly detection benchmarks, 
    OpenOOD \cite{yang2022openood} and ADBench \cite{han2022adbench}. 
    We also show that the benefits of using OOD together with CP apply the other way around by 
    using OOD scores as non-conformity scores, which results in improving upon current CP methods.
    One of the key messages of these contributions is that since OOD is concerned with designing scores and CP with interpreting these scores, the two fields may be inherently intertwined. 
\end{abstract}

  \section{Introduction}
  Machine Learning and Deep Learning models are being increasingly deployed in real-world applications, where they are likely to be confronted with data that is different
  from the data they were trained or validated on. We would like to be able
  to identify this situation, especially for safety-critical Machine Learning applications. More generally, we are interested
  in Out-of-Distribution (OOD) detection, i.e.
  in identifying when an example comes from a particular data distribution for which, in most practical cases,
  we only have access through a dataset of examples drawn from it. 

  Current OOD detection strategies rely on constructing an OOD score $\sood$, 
  a function that assigns a scalar to each input example.
  This score discriminates between in-distribution (ID) data and OOD data by assigning lower scores to the former while assigning higher scores to the latter.

  When OOD detection is used in a machine learning pipeline to identify examples
  that differ from the data the model has been trained on,
  there is a natural qualitative interpretation of OOD detection in terms of uncertainty. 
  For instance, an example with a low OOD score should be one for which the model can 
  predict with low uncertainty, while an example with a high OOD 
  score should be linked to a highly uncertain prediction. 

  Conformal Prediction (CP) is a family of post-hoc methods for Uncertainty Quantification
  that work as wrappers over machine learning models,
  transforming point predictions into prediction sets
  with rigorous probabilistic guarantees based on so-called nonconformity scores.
  The user pre-specifies a risk level $\alpha$, and the constructed prediction set is guaranteed to contain the ground truth value with a probability of at least $1-\alpha$. Since CP is a way of providing rigorous uncertainty quantification guarantees
  built upon scores, it is natural to apply it to the scores used in OOD detection. \textbf{The main purpose of 
  our work is to dig into the Conformal Prediction interpretation of OOD detection scores and show some of its advantages.}

  To that end, we first follow the work of~\cite{bates_testing_2022} on outlier detection
  and apply their ideas to OOD detection. \cite{bates_testing_2022},
  cast the OOD detection problem into the statistical framework of hypothesis testing. They show that the p-values, built with a calibration dataset, are provably marginally valid but depend on the choice of the calibration dataset,
  and so do all the metrics derived from these p-values: FPR, AUROC...
  One of the main contributions of our work is to explore the consequences of 
  this effect for the task of OOD detection and to propose an alternative
  \emph{conformal AUROC} and \emph{conformal FPR} metrics.

  The relevance of the new metrics we propose is best appreciated in the context of safety-critical applications,
  or in an eventual certification process of an OOD detection component.
  The true AUROC or FPR metrics are inaccessible for a given OOD score,
  and we can only provide an approximation obtained from a finite dataset.
  However, this can introduce fluctuations in our approximation,
  thus overestimating or underestimating the true metrics. 
  In a certification process, we are mainly interested in guaranteeing that our estimations are conservative with high probability,
  at the expense of losing some approximation precision.
  Conformal AUROC and Conformal FPR do exactly that. 
  We show the effect of these new metrics on two reference benchmarks,
  the OOD benchmark OpenOOD~\cite{yang2022openood},
  and the anomaly detection benchmark~\cite{han2022adbench}.

  Second, we show that not only can CP contribute to OOD detection, but research in OOD detection
  can help CP too. Indeed, CP has traditionally focused on constructing prediction sets
  from nonconformity scores. Still, the scores used are usually simple functions 
  of the softmax scores for classification tasks or classical distances in Euclidean
  space for regression tasks. Here, we draw inspiration from the OOD detection
  literature to build more involved nonconformity scores and compare
  their performance to the traditional nonconformity scores of CP. For the task of 
  classification, we build prediction sets based on multiple different OOD scores and find that a score based on some of them, notably Mahalanobis \cite{mahalanobis18jesp} or KNN \cite{sun2022knnood} are good candidates as nonconformity scores.

  Ultimately, one of the key messages of these contributions is that since OOD is concerned with designing scores and conformal prediction with interpreting these scores, the two fields may be inherently intertwined. Highlighting this relationship might offer significant potential for cross-fertilization. 

  \subsection*{Summary of Contributions}
  \begin{itemize}
  \item We cast the OOD detection problem into the framework of 
  statistical hypothesis testing and apply the ideas of~\cite{bates_testing_2022} to correct OOD scores.
  \item We propose new conformal AUROC and conformal FPR metrics, which
  are provably conservative with high probability.
  \item We show the effect of conformal AUROC and conformal FPR in the
  reference bechmarks OpenOOD~\cite{yang2022openood} and ADBench~\cite{han2022adbench}.
  \item We build new nonconformity scores for CP based on OOD scores and perform an experimental comparison between the scores. We find
  that the Mahalanobis score outperforms the classical nonconformity score.
  \item We point out that OOD and CP are two domains
  that have much to contribute to each other, 
  and advocate for further research exploring this link.
  \end{itemize}
  

  \section{Background}
  
  \subsection{Out-of-Distribution Detection}\label{sec:background-ood}
  
  Given $n$ examples, $\{ \vx_1, ..., \vx_n\}$ sampled from a probability distribution $\prob{id}$ on a space $\sX$, and a new data point $\vx_{n+1}$, the task of Out-of-Distribution (OOD) detection consists in assessing if $\vx_{n+1}$ was sampled from $\prob{id}$ - in which case it is considered In-Distribution (ID) - or not - thus considered OOD. 
  
  The most common procedure for OOD detection is to construct a score $\sood : \sX \rightarrow \R$ and a threshold $\tau$ such that: 
  \begin{equation}
      \begin{cases}
          \vx_{n+1} \;\;\text{is declared OOD if}\;\; \sood(\vx_{n+1}) > \tau \\
          \vx_{n+1} \;\;\text{is declared ID if}\;\; \sood(\vx_{n+1}) \leq \tau \\
      \end{cases}
  \end{equation}
  We call $\sood$ an OOD score or a non-conformity score.

  \paragraph{Task-based OOD} This is the most common approach in the literature when it comes to OOD detection for neural networks. It also encompasses Open-Set Recognition. 
  Let's consider that $\vx_i$ can be assigned a label $y_i$ so that we can construct a dataset $\{ (\vx_1, y_1),...,(\vx_n, y_n)\}$ defining some supervised deep learning task. In that case, $\prob{id} := \prob{train}$. Task-based OOD uses representations built by the neural network $f$ throughout its training to design $\sood$.  Many sophisticated methods follow this approach  \cite{yang2021generalized}. A simple example is to take the negative maximum of the output softmax of $f$ \cite{hendrycks_baseline_2018} as an OOD score ($\sood(\vx_{n+1}) = -\max \big(f(\vx_{n+1})\big)$ where $\max(\vx)$ is the highest component of the vector $\vx$. Another simple idea is to find the distance to the nearest neighbor in some intermediate layer of $f$ \cite{sun2022knnood}.
  
  \paragraph{Task-agnostic OOD} This approach encompasses One-Class Classification and Anomaly/Outlier Detection. Let's consider a dataset $\{ \vx_1, ..., \vx_n\}$ in a fully unsupervised way. There is no notion of labels, so we have to approximate $\prob{id}$ somehow or some related quantities from scratch. Examples are GANs or VAEs with $\sood$ defined as reconstruction error. See \cite{yang2021generalized} for a thorough review. 
  
  The validation procedure is the same for both approaches. We consider $p$ additional ID samples $\{\vx_{n+1},..., \vx_{n+p}\}$ sampled from $\prob{id}$ (typically, the test set of the corresponding dataset), and $p$ OOD samples $\{\bar{\vx}_1,...,\bar{\vx}_p\}$ from another distribution $\prob{ood} \neq \prob{id} $ (typically, another dataset). We apply $\sood$ to obtain $\{\sood(\bar{\vx}_1),...,\sood(\bar{\vx}_p), \sood(\vx_{n+1}),...,\sood(\vx_{n+p})\}$. Then, we assess the discriminative power of $\sood$ by evaluating metrics depending on the threshold $\tau$. By considering ID samples as negative and OOD as positive, we can compute:
  \begin{itemize}
      \item The Area Under the Receiver Operating Characteristic (AUROC): we compute the False Positive Rate (FPR) and the True Positive Rate (TPR) for $\tau_i = \sood(\vx_{n+i})$, $i \in \{1,...,p\}$, and compute the area under the curve with FPR as x-axis and TPR as y-axis.
      \item FPR@TPR95: The value of the False Positive Rate (FPR) when $\tau$ is selected so that the True Positive Rate (TPR) is $0.95$, i.e. the FPR with $\tau$ such that $1/p \sum_{i} \mathbf{1}_{\sood(\bar{\vx}_i) > \tau} = 0.95$. It can be generalized to FPR@TPR$\beta$, for any $\beta \in (0,1)$.
  \end{itemize}
  \subsection{Conformal Prediction}\label{sec:background-cp}
  Few Machine Learning and Deep Learning models provide a notion of uncertainty related to their predictions.
  Even the models trained for classification tasks providing softmax outputs, which can be interpreted as the probabilities for the input belonging to the different classes,
  are usually ill-calibrated and overconfident, making the softmax output an incorrect proxy of the true uncertainty of the prediction. \cite{pearce2021understanding}.
  Conformal Prediction (CP) \cite{vovk2005algorithmic, Angelopoulos2022} is a series of post-processing uncertainty quantification techniques that are model-agnostic and provide finite-sample guarantees on the model predictions.
  The simplest CP technique, the split CP, works as a wrapper on a trained model $f$. It requires a calibration dataset 
  $\{ (\vx_{n+1}, y_{n+1}),...,(\vx_{n+n_{\text{cal}}}, y_{n+n_{\text{cal}}})\}$ independent of the 
  training data, and a risk (or error rate) $\alpha$ that the user is willing to tolerate. Based on so-called nonconformity scores computed on the calibration dataset, it builds a prediction set $C_{\alpha}(\vx_{n+n_{\text{cal}}+1})$ for a new test sample $\vx_{n+n_{\text{cal}}+1}$ with
  the following finite sample guarantee
  \begin{equation}\label{splitCP}
      P \left( y_{n+n_{\text{cal}}+1} \in C_{\alpha}(\vx_{n+n_{\text{cal}}+1}) \right) \geq 1-\alpha.
  \end{equation}
  In order to obtain the guarantee~\eqref{splitCP}, 
  the only assumption required is that the calibration and test data form an exchangeable sequence (a condition weaker than, and therefore automatically satisfied by independence and identical distribution)\cite{shafer2008tutorial} and
  that they are independent of the training data. 
  It is essential to know that the guarantee~\eqref{splitCP} is marginal, i.e. 
  holds in average over both the choice of the calibration
  dataset and the test sample. As we shall emphasize, there might be fluctuations due to the finite sample size of the calibration dataset.

  
  \section{Related Works}

    \subsection{Statistics Frameworks for OOD}
    In this work, we study the potential of using Conformal Prediction as a statistical framework for interpreting OOD scores. This idea of casting OOD in a statistical framework has already been attempted in different settings.

    \paragraph{Selective Inference and Testing}
    
    Selective Inference works on top of a ML predictor by using an additional decision function 
    in order to decide for each example whether the original model's prediction should be considered. 
    A score equivalent to an OOD score is used to define this decision function. 
    Several approaches exist, for instance, through building a statistical test \cite{haroush_statistical_2022}
    or by training a neural network with an appropriate loss \cite{geifman2017selective,pmlr-v97-geifman19a}. 
    However, the framework of Conformal Prediction appears better suited to our goal since it applies to scores in a post-processing manner, does not require assumptions or modifications on the model, and benefits from dynamic development in the ML community.
    


    \paragraph{Conformal OOD and AD} 
    Conformal Prediction has been previously applied to Out-of-Distribution and Anomaly Detection. For instance, \cite{liang_integrative_2022} have proposed a method based on CP for OOD with labeled outliers, and \cite{kaur_idecode_2022} propose to use conformal p-values. CP is one of several frameworks that allow obtaining statistical guarantees for OOD detection. 
    One of the first methods for Anomaly Detection was introduced by \cite{vovk2003testing}. Since then, several other methods have been proposed by \cite{laxhammar2011sequential,laxhammar2014conformal, balasubramanian2014conformal}, as well as more recently \cite{Angelopoulos2022,guan2022prediction}, where the lengths of the prediction sets as OOD scores. These works all use the standard CP setting, in which basic marginal guarantees are obtained. We go further on this approach by using CP as a probabilistic tool to refine the interpretation and, hence, the usefulness of any OOD score.

    \subsection{Finding Efficient Scores for Conformal Prediction}

    We also investigate the benefits of using OOD scores as non-conformity scores in CP. Common ways to build prediction sets for classification, such as LAC \cite{sadinle2019least} or APS \cite{romano2020classification} and RAPS \cite{angelopoulos2020uncertainty} are based on the softmax output of classifiers. However, non-conformity scores also exist for other predictors \cite{vovk2005algorithmic}, for instance, based on nearest neighbor distance \cite{shafer2008tutorial}. In this work, we suggest interpreting any OOD score as a potential general replacement for scores in CP, opening a large avenue for CP score crafting. This idea could apply to any ML task, but we demonstrate that on a classification task to be consistent with the standard OOD benchmark settings that we follow in the present paper.

 \section{OOD Scores Through the Lens of CP}\label{sec:ood-should-use-cp}

  Let us begin by describing the typical benchmark setup for the evaluation of an OOD score. 
  First, two datasets with non-overlapping labels
  are chosen, one of them  
  to be used as the ID dataset, and the 
  other one, $\dood$ as the OOD dataset. The ID dataset is split into a 
  training set $\didtrain$ and a test set $\didtest$. 
  An OOD detector is trained using the ID training set alone,
  which will output an OOD score for each input provided.
  Finally, the ID test set and the OOD dataset are used
  to evaluate the performance of the obtained OOD score with common evaluation metrics: F-scores, AUROC or
  FPR@TPR$\beta$. 

  \subsection{OOD via Hypothesis Testing}
  Let us first rewrite the problem of OOD detection
  using the framework of statistical hypothesis testing.
  The framework of hypothesis testing allows us to reason in terms of p-values, which have multiple benefits:
  they have a rigorous mathematical definition, and 
  probabilistic interpretation,
  they can be interpreted equivalently for any score,
  and used for comparison of different scores.
  Given a test example $\vx_\text{test}$, we wish to test for
  $\vx_\text{test}\sim\prob{id}$, i.e. we wish to test 
  the null hypothesis $\mathcal{H}_0: \vx_{\text{test}}\sim\prob{id}$
  against the alternate hypothesis $\mathcal{H}_1: \vx_{\text{test}}\not\sim\prob{id}$.
  The value
  $P_{\vx \sim\prob{id}}(\sood(\vx)\geq \sood(\vx_{\text{test}}))$
  is an exact p-value for the null hypothesis $\mathcal{H}_0$. Since we don't have access to 
  the distribution $\prob{id}$, 
  a first naive approach would be to approximate 
  $\prob{id}$ with the empirical distribution 
  obtained from a sample
  of i.i.d. random variables $\vx_i\sim\prob{id}$,
  we call this sample the calibration set, and we denote
  it by $\didcal$.
  However, fluctuations in this sample can lead to 
  over-confident estimations of the p-value. In order
  to avoid obtaining over-confident estimations of the 
  p-value,
  we use the correction proposed by~\cite{bates_testing_2022}
  (which can be originally traced to~\cite{papadopoulos2002inductive}):
  \begin{equation}
  \umarg(\vx)=
  \frac{1+|\lbrace i\in\didcal\,:\, \sood(\vx)\geq \sood(\vx_i)\rbrace|}{1+n_\text{cal}}.
  \end{equation}
  With this correction, we obtain \emph{marginally valid}
  p-values, that is, p-values that satisfy
  \begin{equation}
  P_{\vx\sim\prob{id}}\big(
  \umarg(\vx) \leq t
  \big)\leq t,\quad \text{for all}\quad 0\leq t \leq 1.
  \end{equation}
  By \emph{marginally}, we are pointing out that the 
  probability in the above formula integrates over both
  the calibration set $\didcal$ and the test point $\vx$.
  However, for a given calibration set, the corrected p-value
  may still be overly confident, as we show in the next section. 

  \subsection{Fluctuations of the p-value}\label{sec:fpr-as-a-random-variable}

%
%
In fact, \cite{bates_testing_2022} points out that the quantity 
    $$F(\tau;\didcal):=
    P_{\vx\sim\prob{id}}
    \left(\umarg(\vx)\leq\tau\,|\,\didcal\right).$$
%
%
%

  \begin{figure}[h]
  \centering
  \includegraphics[width=0.5\textwidth]{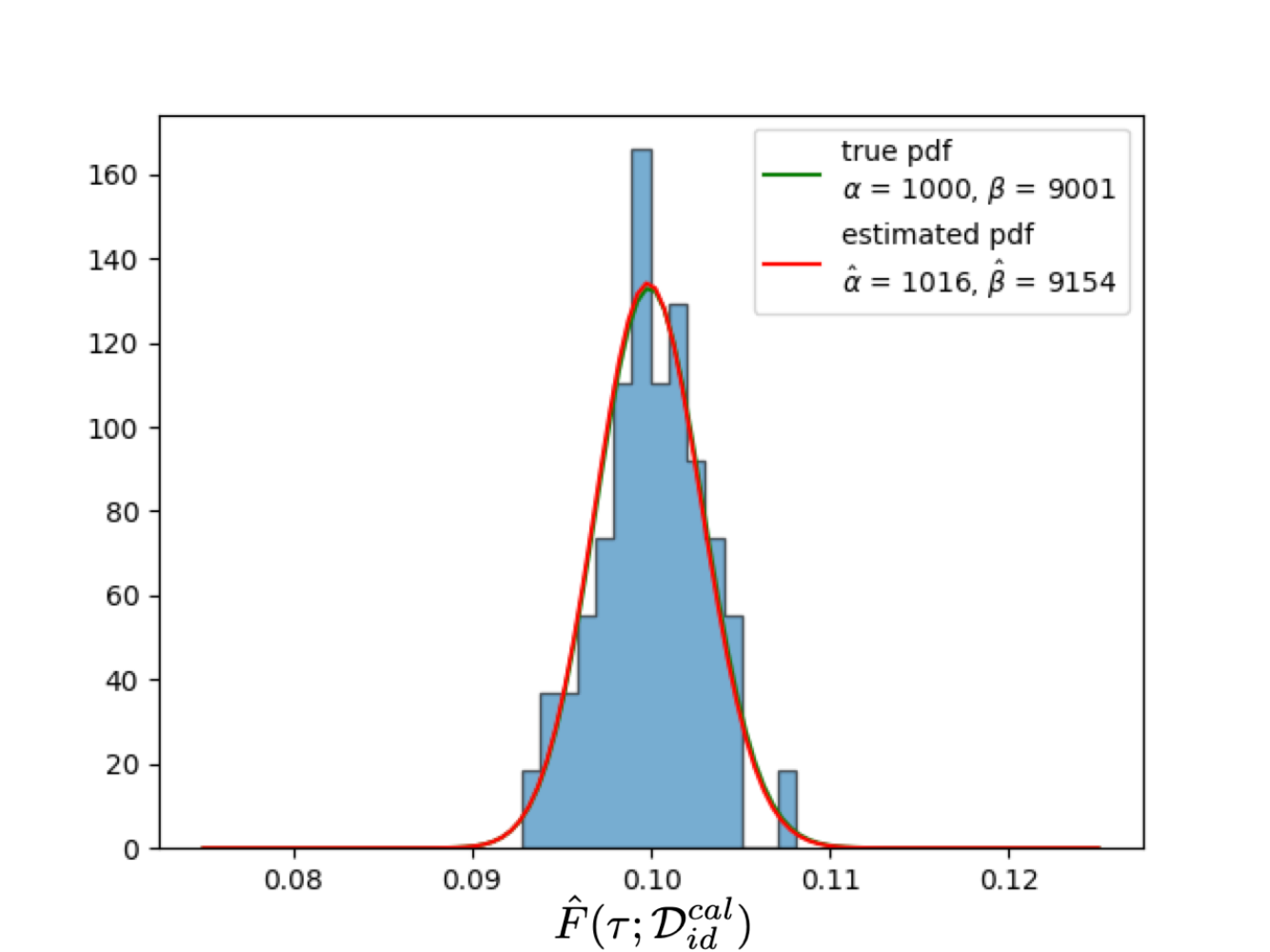}
  \caption{Histogram of $F(0.1;\didcal)$ for different calibration sets. 
  The histogram is obtained by splitting the dataset \emph{svhn\_extra} 
  into disjoint calibration sets of $10000$ points each, 
  and approximating the value of $F$ for each calibration set 
  by integrating over the remaining 521131 examples.}
  \label{fig:beta}
  \end{figure}

           \begin{figure*}[!h]
  \label{fig:ccpv}
  \centering
  \includegraphics[width=1.0\textwidth]{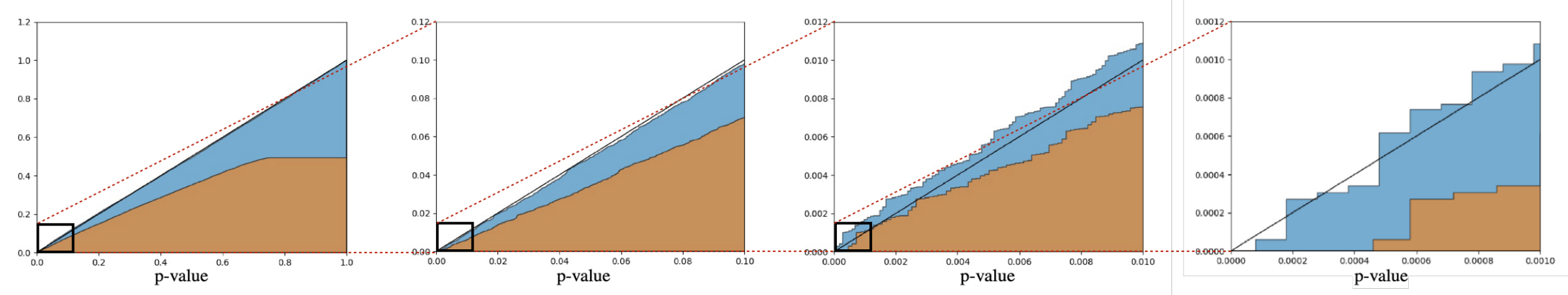}
  \vspace{-0.7cm}
    \caption{Cumulative histogram of the marginal p-values (blue) and the calibration-conditional p-values (brown) obtained by performing the Simes adjustment method. Four zooms of the same plot are shown, obtained with a calibration dataset of 10000 points from the SVHN dataset. The approximation of the marginal p-values becomes poor for smaller values of $\alpha$, and it can be overly optimistic. The correction is conservative for all values of $\alpha$ \emph{simultaneously}, as shown in the figure, which happens with probability $\delta=0.1$ over the choice of the calibration dataset.}
  \end{figure*}
 is a random variable that depends on the 
    calibration set $\didcal$. As a practical consequence, the same OOD score $\sood$ 
    corrected using two different calibration datasets will give rise to 
    different $p$-values.  
    The random variable $F(\tau;\didcal)$ 
    follows a distribution that is known: 
    it is a Beta distribution that depends on the parameters $n_{cal}$ and $\tau$:
    \begin{equation}
    \label{eq:beta}
            F(\tau;\didcal)\sim \text{Beta}(\ell, n_{cal}+1-\ell),
    \end{equation}
    where $\ell=\lfloor (n_{cal}+1)\tau\rfloor$ (cf. \cite{bates_testing_2022} or \cite{vovk_conditional_2012} for a proof of the result). 
    In order to illustrate this phenomenon, we leverage the fact that SVHN dataset provides an additional set of 530000 \emph{extra} test images. It allows simulating 53 draws of the 
    random variable $F(\tau;\didcal)$, by splitting the over 530000 examples in the 
    \emph{svhn\_extra} dataset into 53 different folds of 10000 examples each.  
    For each fold, the 10000 examples are used to constitute the calibration dataset
    $\didcal$, whereas the remaining over 520000 examples are used to approximate
    the computation of $F$, i.e., given a calibration dataset $\didcal$,
    \begin{equation}
        F(\tau;\didcal) \approx \hat{F}(\tau;\didcal) = \frac{1}{520000}\sum_{k=1}^{520000} 1_{\umarg(\vx_k)\leq\tau}.
    \end{equation}
    Due to the large number of points used in the approximating sum,
    the 53 values obtained are faithful approximations of the random variables
    $F(\tau;\didcal)$. 
    We perform this simulation with $\tau=0.1$ and plot the 53 values into a histogram. 
    Additionally, we fit
    a Beta distribution to the histogram using the \emph{scikit-learn} library. 
    These plots are found in figure~\ref{fig:beta}.
    As we can see, the estimated parameters of the fitted beta 
    distribution are very close to those predicted by 
    the theoretical result of~\eqref{eq:beta}.
    If the value $\umarg(\vx)$ were a true p-value, the value of $F(\tau;\didcal)$
    would be equal to $\tau$,
    but as we can see from the theoretical result and the experiment above, 
    $F(\tau;\didcal)$ is a random variable that fluctuates around its 
    mean value $\tau$.

  \subsection{Probabilistic Guarantees for the p-values and FPR}\label{sec:probabilistic-guarantees-for-the-fpr}

    In order to have guaranteed conservative estimates for the p-values,
    we follow~\cite{bates_testing_2022} in further correcting the marginal 
    p-values thus obtaining \emph{calibration-conditional} p-values.
    Given a user-predefined risk level $\delta$,
    the calibration-conditional p-values $\ucc$ will satisfy
    \begin{equation}
    \label{eq:ccpvals}
  P\Big(P
  \big(
  \ucc(\vx) \leq t \,|\, \didcal
  \big)\leq t,\ \forall\, t\in(0,1)\Big)\geq 1-\delta,
  \end{equation}
  where the probability inside is taken over $\vx\sim\prob{id}$,
  and the probability outside over the choice of $\didcal$.
  Thus, with a probability of at least $1-\delta$, we can be confident that
  we have a \emph{good} calibration set, meaning that our p-values will be
  conservative. An illustration of the difference between marginal p-values
  and calibration-conditional p-values is given in~\ref{fig:ccpv},
  where the cumulative histograms of both are compared using the so-called
  Simes correction for obtaining the calibration-conditional p-values.

Likewise, we can correct the FPR directly.
We denote by $\fpr(\tau)$ the true FPR and by $\widehat{\fpr}_n(\tau)$ 
the empirical approximation of the FPR obtained using $n$ samples, i.e.,
\begin{align}
    \fpr(\tau) &= P_{\vx\sim\prob{id}}(\sood(\vx)\geq\tau),\\
    \widehat{\fpr}(\tau) &= \frac{1}{n_{\text{cal}}}\sum_{\vx_i\in\didcal}1_{\sood(\vx_i)\geq \tau}.
\end{align}
Note that the FPR and the p-values are intrinsically linked -- we have that $\hat{u}^{\text{marg}}(\vx) = (n_{\text{cal}}\widehat{\fpr}_n(\sood(\vx)) + 1)/(1 + n_{\text{cal}})$) -- so as for the p-values, the empirical approximation $\widehat{\fpr}(\tau)$ also depends on the calibration dataset $\didcal$
and can give rise to overly confident estimations of the true FPR. This effect can impact other metrics defined based on this empirical
FPR. In order to obtain high-confidence conservative estimates of the FPR, 
\cite{bates_testing_2022} propose a correction of the empirical FPR that satisfies the following:
  \begin{equation} 
  \label{eq:uni_fpr}
\mathbb{P}\left[\fpr(\tau) \leq \widehat{\fpr}^{+}(\tau), \forall \tau \in \mathbb{R}\right] \geq 1-\delta,
\end{equation}
where $\widehat{\fpr}^{+}(\tau)$ is a correction version of the empirical $\widehat{\fpr}(\tau)$. 
The corrected FPR is obtained by applying a correction function $h$
to the empirical FPR, i.e.
$\widehat{\fpr}^{+}(\tau)=h\circ\widehat{\fpr}(\tau)$.

The calibration-conditional p-values are obtained by applying a correction function $h$
the marginal p-values, i.e. 
$\ucc (\vx) = h\circ \umarg(\vx)$.

  Four different correction functions $h$ are proposed
  by~\cite{bates_testing_2022}, the Simes, DKWM, Asymptotic
  and Monte Carlo corrections. The Simes, DKWM
 and Monte Carlo corrections all provide the finite sample
  guarantees of~\eqref{eq:ccpvals} and~\eqref{eq:uni_fpr}, while the 
  Asymptotic correction provides only an asymptotic guarantee,
  that is, when the number of calibration points goes to
  infinity. Between the three corrections providing the finite sample guarantee,
  we find the Monte Carlo one to give tighter bounds 
  (cf. the appendix~\ref{app:SimesMC} for more details on how the Simes and Monte Carlo corrections are defined).

        \begin{figure*}[!t]
  \label{fig:roc}
  \centering
  \includegraphics[width=1.0\textwidth]{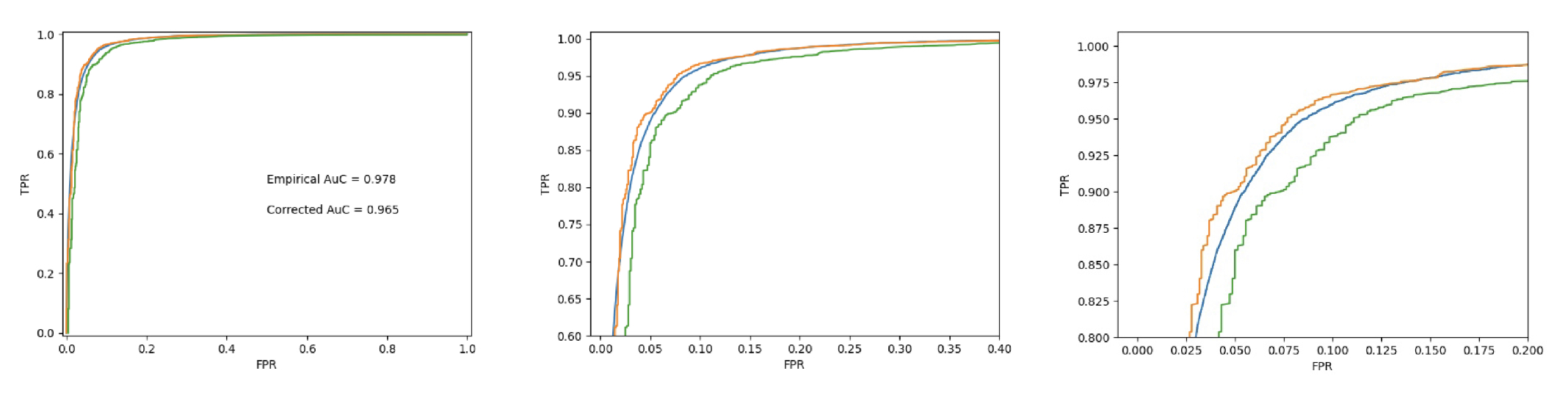}
  \vspace{-0.7cm}
    \caption{Different zoom levels of the ROC curves. The TPR is calculated by using all the points in the "Cifar10" dataset for the three curves. As for the TPR, the blue curve is obtained by using all data points in the "svhn\_extra" dataset, the orange curve is an approximation of the blue curve using 1000 calibration points, whereas the green curve is obtained by correcting the FPR via the conformal AUROC method.}
  \end{figure*}  

\subsection{Conformal Metrics for OOD}
    Based on the above discussion, 
    we define the notions of conformal FPR and then conformal AUROC and FPR@TPR95 that are built upon conformal FPR.
    The conformal FPR is obtained by applying the correction function $h$ to the empirical FPR
    as explained in the previous section,
    while the conformal AUROC is obtained by plotting 
    the AUROC curve having conformal FPR as the $x$-axis. Similarly, conformal FPR@TPR95 is obtained by computing the classical FPR@TPR95 and then applying the correction. The computations are performed by considering the ID validation dataset as the calibration dataset.
    Conformal FPR, AUROC, and FPR@TPR95 are not 
    necessarily better approximations of the real FPR, 
    AUROC and FPR@TPR95 values. Nonetheless, they are guaranteed to use 
    conservative estimates of the FPR with a user-defined miscalibration tolerance
    $\delta$, which is an essential property in many safety-critical
    applications or certification processes \cite{Sellke2001CalibrationO}.
    The effect of the correction on the ROC curve is illustrated in Figure 
    \ref{fig:roc} using the SVHN dataset as ID and Cifar-10 as OOD. 

  \subsection{Safer Benchmarks for OOD}\label{sec:benchmarks}

  AUROC and (to a lesser extent) FPR@TPR$95$ are two metrics that OOD and AD practitioners intensively use to benchmark and evaluate the performances of different OOD  detection algorithms. However, as we saw in the previous sections, the evaluation can be overly optimistic, which can be detrimental to algorithms designed for safety-critical applications. In this section, we reevaluate various OOD baselines included in the very furnished OpenOOD \cite{yang2022openood}, and ADBench \cite{han2022adbench} benchmarks and illustrate the trade-off between performances and probabilistic guarantees.  
  
  \subsubsection{OpenOOD}\label{sec:openood}
  
   OpenOOD \cite{yang2022openood} is an extensive benchmark for task-based OOD, i.e. for OOD methods that assess if some test data resembles some trained backbone's training data. Usually, backbones trained on CIFAR-10, CIFAR-100, Imagenet200, and Imagenet are considered. In our case, we consider a ResNet18 trained on the first three datasets only since we are not evaluating a new baseline but only investigating a new metric for the benchmark. We evaluate the AUROC of several baselines with various OOD datasets gathered into two groups, Near OOD and Far OOD, following OpenOOD's guidelines. We then compute the correction for the AUROC, with $\delta=0.01$. The results are displayed in Table \ref{tab:openood}. We also run the benchmark for $\delta=0.05$ and FPR-$95$, which we defer to Appendix \ref{app:appB}.
   
  \begin{table*}[ht]
    \centering
    \resizebox{\textwidth}{!}{
    \begin{tabular}{l|cccc|cccc|cccc} 
        \toprule
         & \multicolumn{4}{c|}{\textbf{CIFAR-10}} & \multicolumn{4}{c|}{\textbf{CIFAR-100}} & \multicolumn{4}{c}{\textbf{ImageNet-200}} \\
        
        \midrule
        OOD type  & \multicolumn{2}{c}{Near OOD} & \multicolumn{2}{c|}{Far OOD}  & \multicolumn{2}{c}{Near OOD} & \multicolumn{2}{c|}{Far OOD}  & \multicolumn{2}{c}{Near OOD} & \multicolumn{2}{c}{Far OOD}  \\
           & class. & conf. & class. & conf. & class. & conf. & class. & conf. & class. & conf. & class. & conf. \\
        \midrule

        OpenMax \cite{openmax16cvpr} & {87.2} & \textbf{85.95} & {89.53} & \textbf{88.3} & {76.66} & \textbf{74.95} & {79.12} & \textbf{77.52} & {80.4} & \textbf{78.82} & {90.41} & \textbf{88.77} \\ 
        
        MSP \cite{hendrycks_baseline_2018} & {87.68} & \textbf{86.56} & {91.0} & \textbf{89.98} & {80.42} & \textbf{78.93} & {77.58} & \textbf{76.0} & {83.3} & \textbf{81.85} & {90.2} & \textbf{88.83} \\ 
        
        TempScale \cite{guo2017calibration} & {87.65} & \textbf{86.55} & {91.27} & \textbf{90.3} & {80.98} & \textbf{79.51} & {78.51} & \textbf{76.95} & {83.66} & \textbf{82.21} & {90.91} & \textbf{89.53} \\ 
        
        ODIN \cite{odin18iclr} & {80.25} & \textbf{79.04} & {87.21} & \textbf{86.26} & {79.8} & \textbf{78.3} & {79.44} & \textbf{77.92} & {80.32} & \textbf{78.85} & {91.89} & \textbf{90.59} \\ 
        
        MDS \cite{mahananobis18nips} & {86.72} & \textbf{85.49} & {90.2} & \textbf{89.09} & {58.79} & \textbf{56.85} & {70.06} & \textbf{68.31} & {62.51} & \textbf{60.68} & {74.94} & \textbf{73.09} \\ 
        
        MDSEns \cite{mahananobis18nips} & {60.46} & \textbf{58.69} & {74.07} & \textbf{72.72} & {45.98} & \textbf{43.97} & {66.03} & \textbf{64.43} & {54.58} & \textbf{52.76} & {70.08} & \textbf{68.35} \\ 
        
        Gram \cite{gram20icml} & {52.63} & \textbf{50.69} & {69.74} & \textbf{68.11} & {50.69} & \textbf{48.69} & {73.97} & \textbf{72.63} & {68.36} & \textbf{66.74} & {70.94} & \textbf{69.3} \\ 
        
        EBO \cite{energyood20nips} & {86.93} & \textbf{85.9} & {91.74} & \textbf{90.9} & {80.84} & \textbf{79.36} & {79.71} & \textbf{78.19} & {82.57} & \textbf{81.1} & {91.12} & \textbf{89.71} \\ 
        
        GradNorm \cite{huang2021importance} & {53.77} & \textbf{51.92} & {58.55} & \textbf{56.76} & {69.73} & \textbf{68.11} & {68.82} & \textbf{67.19} & {73.33} & \textbf{71.85} & {85.29} & \textbf{83.99} \\ 
        
        ReAct \cite{react21nips} & {86.47} & \textbf{85.41} & {91.02} & \textbf{90.12} & {80.7} & \textbf{79.23} & {79.84} & \textbf{78.32} & {80.48} & \textbf{79.0} & {93.1} & \textbf{91.79} \\ 
        
        MLS \cite{species22icml} & {86.86} & \textbf{85.81} & {91.61} & \textbf{90.74} & {81.04} & \textbf{79.58} & {79.6} & \textbf{78.07} & {82.96} & \textbf{81.5} & {91.34} & \textbf{89.94} \\ 
        
        KLM \cite{species22icml}  & {78.8} & \textbf{77.58} & {82.76} & \textbf{81.63} & {76.9} & \textbf{75.38} & {76.03} & \textbf{74.52} & {80.69} & \textbf{79.14} & {88.41} & \textbf{86.74} \\ 
        
        VIM \cite{haoqi2022vim}  & {88.51} & \textbf{87.42} & {93.14} & \textbf{92.25} & {74.83} & \textbf{73.17} & {82.11} & \textbf{80.69} & {78.81} & \textbf{77.2} & {91.52} & \textbf{90.05} \\ 
        
        KNN \cite{sun2022knnood} & {90.7} & \textbf{89.69} & {93.1} & \textbf{92.19} & {80.25} & \textbf{78.79} & {82.32} & \textbf{80.93} & {81.75} & \textbf{80.27} & {93.47} & \textbf{92.25} \\ 
        
        DICE \cite{sun2021dice} & {77.79} & \textbf{76.44} & {85.41} & \textbf{84.37} & {79.15} & \textbf{77.61} & {79.84} & \textbf{78.33} & {81.97} & \textbf{80.5} & {91.19} & \textbf{89.84} \\ 
        
        RankFeat \cite{song2022rankfeat} & {76.33} & \textbf{74.76} & {70.15} & \textbf{68.39} & {62.22} & \textbf{60.33} & {67.74} & \textbf{65.9} & {58.57} & \textbf{57.0} & {38.97} & \textbf{37.09} \\ 
        
        ASH \cite{djurisic2023extremely} & {74.11} & \textbf{72.71} & {78.36} & \textbf{77.02} & {78.39} & \textbf{76.89} & {79.7} & \textbf{78.23} & {82.12} & \textbf{80.72} & {94.23} & \textbf{93.11} \\ 
        
        SHE \cite{she23iclr} & {80.84} & \textbf{79.64} & {86.55} & \textbf{85.55} & {78.72} & \textbf{77.18} & {77.35} & \textbf{75.8} & {80.46} & \textbf{79.0} & {90.48} & \textbf{89.17} \\

        \bottomrule
        \end{tabular}
    }
    \caption{Classical AUROC (class.) vs Conformal AUROC (conf.) obtained with the Monte Carlo method and $\delta=0.01$ for several baselines from OpenOOD benchmark.}
    \label{tab:openood}
    \end{table*}

  Table \ref{tab:openood} shows that after the correction, the conformal AUROC is lower than the classical AUROC, by often more than $1$ percent. On the one hand,  this is significant, especially for such benchmarks where the State-of-the-art often holds by a fraction of a percentage. On the other hand, the correction is not \textit{so} severe, and the best baselines still get very good AUROC despite the correction. In other words, the correction is large enough to manifest its importance but low enough to still be useable in practice: \textit{it costs only roughly $1$ or $2$ percent in AUROC to be $99\%$ sure that the FPR involved in the AUROC calculation is not overestimated.}
  
  \subsubsection{ADBench}\label{sec:adbench}

\begin{figure*}[!h]
        \centering
        \includegraphics[width=0.35\linewidth,trim={0cm 0cm 0cm 0cm},clip]{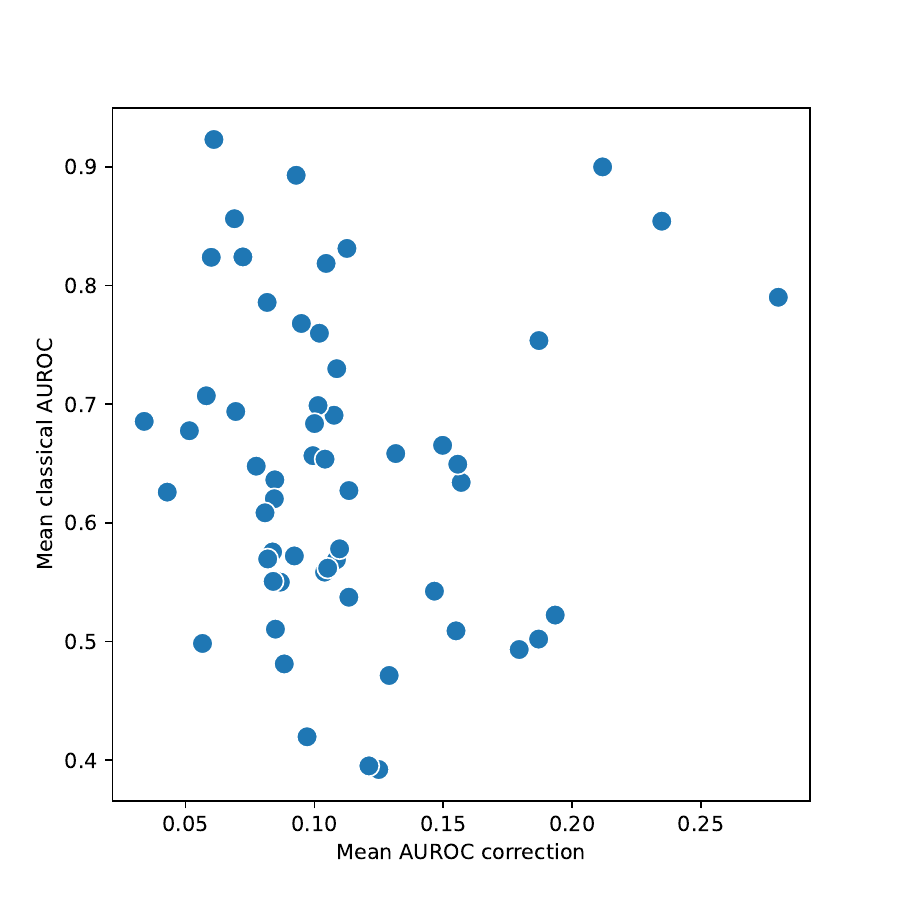}
        \includegraphics[width=0.64\linewidth,trim={2cm 0cm 3cm 0cm},clip]{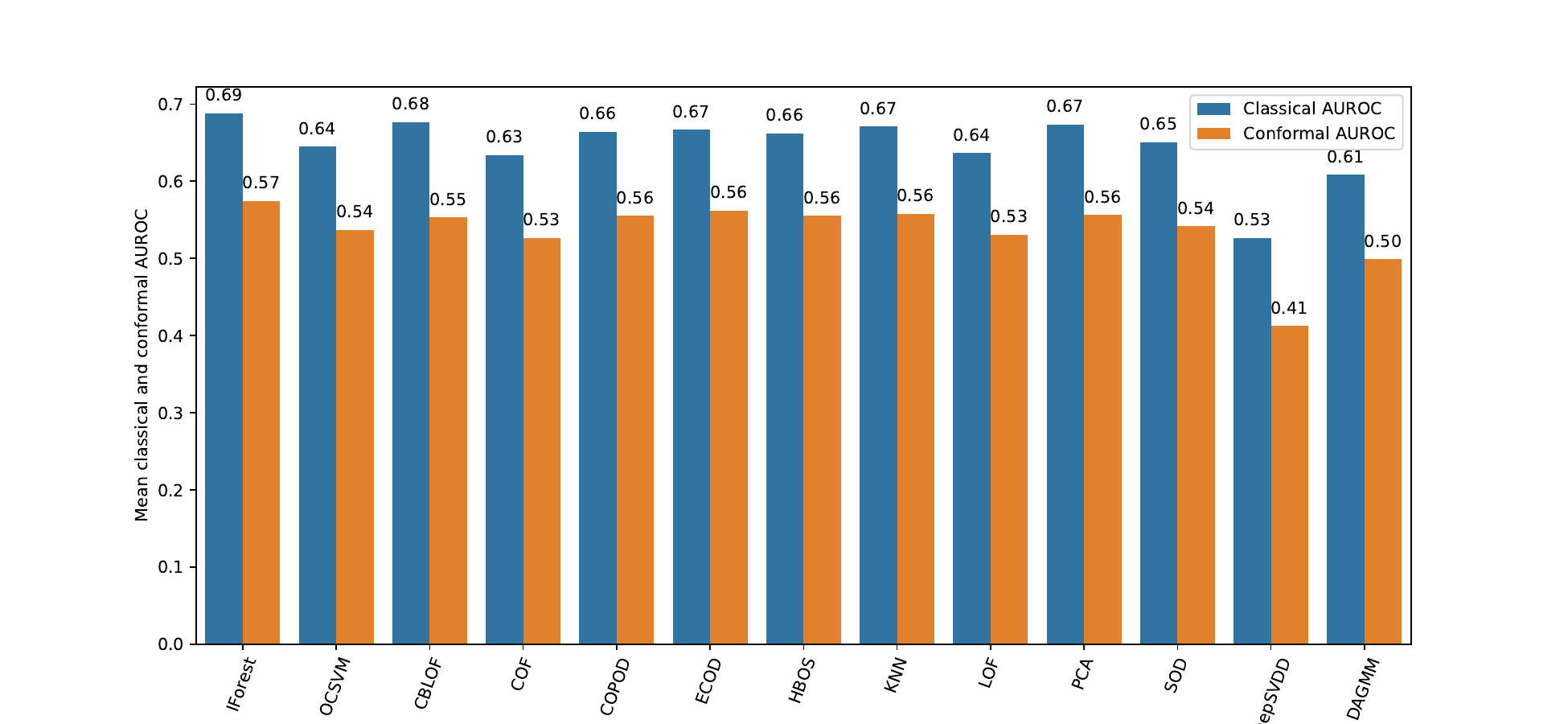}
        \caption{Results visualization for ADBench benchmark. \textbf{(left)} Scatter plot with mean classical AUROC and mean AUROC correction over different methods for each dataset as y-axis and x-axis, respectively. \textbf{(right)} Mean AUROC and AUROC correction over different datasets for each AD method.}
        \label{fig:adb1}
  \end{figure*}

  We perform the same procedure as OpenOOD with ADBench \cite{han2022adbench}, which gathers many task-agnostic OOD baselines -- considered Anomaly Detection (AD), hence the benchmark's name. We conduct the experiments with "unsupervised AD" baselines, i.e. baselines that do not leverage labeled anomalies. We apply the correction with $\delta=0.05$ and summarize the results in Figure \ref{fig:adb1}. 
  The complete results are deferred to Appendix \ref{app:appC}.
  
  Figure \ref{fig:adb1} (left) shows a scatter plot with mean classical AUROC and mean AUROC correction over different methods for each dataset as y-axis and x-axis, respectively. The variability and magnitude of the correction are higher than for OpenOOD since the number of points in the test set changes depending on the dataset and is generally way lower. This observation is important because it illustrates the brittleness of the conclusions that can be drawn from AD benchmarks and supports the increasingly commonly accepted fact that no method is provably better than others in AD -- one of the key conclusions of ADBench's paper itself \cite{han2022adbench}. Figure \ref{fig:adb1} (right) shows the mean classical and conformal AUROC for each baseline over the datasets. The correction is more stable, demonstrating that the correction affects all baselines similarly. 
  
  \section{A Deeper Link between CP and OOD}\label{sec:deeper-link}
  
  In the previous sections, we have mostly emphasized that practitioners of OOD detection should look at CP as an additional building block for correctly interpreting the scores that all the OOD methods rely on. In this section, we advocate that the link between OOD and CP goes even deeper and that both fields could benefit from each other.
\label{sec:cp-should-use-ood}

    \begin{table*}[!t]
      \centering
      \resizebox{\textwidth}{!}{   \label{tab:cp-scores}
      \begin{tabular}{l|ccc|ccc|ccc}
          \toprule
                      &  \multicolumn{3}{c|}{\textbf{LAC}}  &  \multicolumn{3}{c|}{\textbf{APS}} & \multicolumn{3}{c}{\textbf{RAPS}} \\
          \midrule
          $\alpha$     & $0.005$ & $0.01$ & $0.05$  & $0.005$ & $0.01$ & $0.05$ & $0.005$ & $0.01$ & $0.05$ \\
          \midrule
          Gram         & $9.57$ & $8.34$ & $1.89$ & $9.60 \pm 0.10$ & $1.93 \pm 0.06$ & $8.66 \pm 0.13$ & $9.56 \pm 0.13$ & $8.7 \pm 0.16$ & $1.89 \pm 0.03$ \\
          ReAct        & $3.75$ & $1.98$ & $1.03$ & $4.47 \pm 0.16$ & $1.97 \pm 0.09$ & $3.62 \pm 0.17$ & $4.46 \pm 0.15$ & $3.67 \pm 0.19$& $2.02 \pm 0.09$  \\
          ODIN         & $7.15$ & $5.82$ & $1.14$ & $7.42 \pm 0.17$ & $1.53 \pm 0.06$ & $5.14 \pm 0.08$ & $7.45 \pm 0.16$ & $5.1 \pm 0.10$ & $1.57 \pm 0.08$ \\
          KNN         & $2.57$ & $\textbf{1.48}$ & $\textbf{1.01}$ & $\underline{3.62} \pm 0.15$ & $\underline{1.09} \pm 0.03$ & $2.71 \pm 0.11$ & $\underline{3.69} \pm 0.11$ & $2.77 \pm 0.09$ & $\underline{1.08} \pm 0.02$  \\
          Mahalanobis  & $\textbf{1.85}$ & $\underline{1.47}$ & $1.04$ & $\textbf{1.89} \pm 0.07$ & $\textbf{1.04} \pm 0.01$ & $\textbf{1.49} \pm 0.04$ & $\textbf{1.92} \pm 0.05$ & $\textbf{1.49} \pm 0.05$& $\textbf{1.04} \pm 0.01$  \\
          Softmax           & $\underline{2.44}$ & $1.73$ & $\underline{1.03}$ & $3.92 \pm 0.26$ & $1.1 \pm 0.01$ & $\underline{2.16} \pm 0.13$ & $3.81 \pm 0.24$ & $\underline{2.17} \pm 0.11$& $1.09 \pm 0.01$  \\
          \bottomrule
      \end{tabular}
      }
      \caption{Efficiency (mean $\pm$ standard deviation for APS and RAPS) of the prediction sets for different scores for Conformal classification on Cifar10. The best is bolded, and the second is underlined.}
   
  \end{table*}

  So far, we have shown how OOD can use CP, but we argue that CP could also use OOD. Indeed, CP is about interpreting scores to provide probabilistic results. But CP works regardless of the given score. Indeed, all scores will have the same guarantee, but better scores will give tighter prediction sets, and 
  worse scores will give very large and uninformative prediction sets. For CP to provide powerful probabilistic guarantees, the scores have to be informative, hence the common practice of relying on scores derived from the softmax values of a neural network
  \cite{sadinle2019least}. It turns out that the softmax is also a score used in OOD detection \cite{hendrycks_baseline_2018}, which suggests that OOD scores and CP scores might be related in some way. In this section, we explore using different OOD scores to perform CP. We consider a ResNet18 trained on CIFAR-10 and build conformal prediction sets following the procedure described in \secref{sec:background-cp}. To build these prediction sets, we use scores based on ReAct \cite{react21nips}, Gram \cite{gram20icml}, KNN \cite{sun2022knnood}, Mahalanobis \cite{mahananobis18nips}, and ODIN \cite{odin18iclr}. Note that we had to adapt those scores to make them class-dependent since the score used in CP is defined as $\scp(\vx, y)$. We did so using the OOD library \href{https://github.com/deel-ai/oodeel}{oodeel}\footnote{https://github.com/deel-ai/oodeel}, following a procedure that we describe in detail in Appendix \ref{app:appA}. Then, given the OOD score $\sood(\vx, y_i)$, we construct softmax-like scores $\hat{s}(\vx, y_i) = \exp{\sood(\vx, y_i)} / \sum_j \exp{\sood(\vx, y_j)}$ , and use it for CP.

  For each defined score, we perform the calibration step on $n_{cal} = 2000$ points following the classical Least-Ambiguous set classifiers (LAC) procedure \cite{sadinle2019least}, and the more recent Adaptive Prediction Set (APS) \cite{romano2020classification} and Regularized Adaptive Prediction Set (RAPS) \cite{angelopoulos2020uncertainty} methods. For all methods, we construct the prediction sets for each of the remaining $n_{val} - n_{cal} = 8000$ points, and for coverages $1 - \alpha \in \{0.005, 0.01, 0.05\}$. The prediction sets calculations are implemented using the library \href{https://github.com/deel-ai/puncc}{PUNCC}\footnote{https://github.com/deel-ai/puncc}. We assess the mean efficiency of the prediction sets for each score, including LAC, APS, and RAPS based on softmax, as classically done in CP in Table \ref{tab:cp-scores}). Since APS and RAPS involve sampling a uniform random variable, we report the mean and the standard deviation of the mean efficiency for $10$ evaluations. 
  
  Table \ref{tab:cp-scores} shows that all OOD scores are inefficient for CP. For example, Gram performs very poorly. However, in some instances, some scores, like KNN or Mahalanobis, perform better than classical CP scores. This suggests that OOD scores may be good candidates as nonconformity scores.
  
We used a limited number of OOD scores in our experiment, but many more are available. This section highlights the need to explore nonconformity score crafting in this research area rather than demonstrate the superiority of Mahalanobis or KNN over nonconformity scores.
  
  

  

  \vspace{-0.2cm}
\section{Conclusion \& Discussion}

In conclusion, our work highlights the inherent randomness of OOD metrics and demonstrates how Conformal Prediction (CP) can effectively correct these metrics. We have also shown that recent advancements in CP allow for uniform conservativeness guarantees on OOD metrics, providing more reliable evaluations. Furthermore, our analysis reveals that the correction introduced by CP does not significantly impact the performance of the best OOD baselines. On the other hand, we also showed that we could use OOD to improve existing CP techniques by using OOD scores as nonconformity scores. We found that some of them, especially Mahalanobis and KNN, are good candidates for nonconformity scores, unlocking a whole avenue for crafting CP nonconformity scores based on the plethora of existing post-hoc OOD scores.  

By integrating CP with OOD, we have demonstrated the fruitful synergy between the two fields. OOD detection focuses on developing scores that accurately discriminate between OOD and ID, while CP specializes in interpreting scores to provide probabilistic guarantees. This interplay between OOD and CP presents opportunities for mutual advancement: advancements in CP research can enhance OOD by offering more refined probabilistic interpretations of OOD scores, which is particularly crucial in safety-critical applications. Conversely, progress in OOD research can benefit CP by providing scores that improve the efficiency of prediction sets. This suggests that further exploration and collaboration between the two fields hold great potential.

In summary, our findings underscore the intertwined nature of OOD and CP, emphasizing the need for continued investigation and cross-fertilization to advance both disciplines.

\newpage
\section*{Impact statement}
This paper presents work whose goal is to advance the field of Machine Learning. There are many potential societal consequences of our work, none which we feel must be specifically highlighted here.
\bibliographystyle{icml2024}
\bibliography{refs}

\appendix
\onecolumn
\section{Appendix: Simes and Monte Carlo corrections}
\label{app:SimesMC}

In our work we use two of the corrections proposed by \cite{bates_testing_2022}, Simes and Monte Carlo Correction. 
In this section, we introduce these corrections for the sake of completeness, as well as two other corrections that we do not use for reasons to be detailed. 

\paragraph{Simes Correction}
   Generally, we are interested in small p-values and the Simes correction focuses on those, that is by adding a smaller correction to the smaller p-values than the larger ones.
   \begin{equation}
   b_{n+1-i}^{\mathrm{s}}=1-\delta^{2/n}\left(\frac{i \cdots(i-n/2+1)}{n \cdots(n-n/2+1)}\right)^{2/n}, \quad i=1, \ldots, n
   \end{equation}

   \paragraph{DKWM}
   The former approach may be compared to the classical uniform concentration DKWM result, where the $b$ are defined as
   \begin{equation}
b_i^{\mathrm{d}}=\min \{(i / n)+\sqrt{\log (2 / \delta) / 2 n}, 1\} \text {; }
\end{equation}
   However, DKWM tends to provide much larger bounds than Simes.

   \paragraph{Asymptotic Correction}
   The previous approach brought finite sample guarantees but at the cost of a large correction. In order to produce a tighter bound, for a more powerful test, we look into a correction that is correct asymptotically.
   \begin{equation}
   \begin{aligned}
c_n(\delta)&:=\left(\sqrt{2 \log \log n}\right)^{-1}\left(-\log [-\log (1-\delta)]\right. \\ & \left. +2 \log \log n+(1 / 2) \log \log \log n-(1 / 2) \log \pi\right).
   \end{aligned}
   \end{equation}
   \begin{equation}
b_i^{\mathrm{a}}=\min \left\{\frac{i}{n}+c_n(\delta) \frac{\sqrt{i(n-i)}}{n \sqrt{n}}, 1\right\}, \quad i=1, \ldots, n
\end{equation}
   This bound is quite similar to Simes for small values, but quite tighter for the remaining ones.

   \paragraph{Monte Carlo Correction}
   The Monter Carlo Correction offers advantages of both the Simes and Asymptotic methods. It provides a finite-sample guarantee, mimics Simes for small p-values and remains closer to the asymptotic correction for larger ones.
   \begin{equation}
h^{\mathrm{m}, \hat{\delta}}(t)=\min \left\{h^{\mathrm{s}}(t), h^{\mathrm{a}, \hat{\delta}}(t)\right\}, \quad t \in[0,1] .
\end{equation}

\section{Appendix: Designing class-dependent OOD scores for CP}
\label{app:appA}

Let's consider a classification task with a classifier $f$ trained to fit a dataset $\{(\vx_1, y_1),...,(\vx_n, y_n)\}$, where $\vx_i \in \sX$ and $y_i \in \{1,...C\}$ for all $i \in \{1,...,n\}$. In OOD, the score function $\sood: \sX \rightarrow \R $, whereas in CP, the non-conformity score $\scp: \sX \times \R \rightarrow \R$. Hence, in order to construct a non-conformity score out of $\sood$, we have to make it class-dependent. In this section, we describe how to construct class-dependent OOD scores out of classical OOD scores for appropriate usage in CP.

\subsection{ReAct}

ReAct method \cite{react21nips} gets the quantiles of $f$'s penultimate layer's activation values and then clips the activation values for a new input data point. The output softmax are then used for OOD scoring. Therefore, making the score class-dependent is straightforward: one only has to get the class softmax.

\subsection{ODIN}

The idea of ODIN \cite{odin18iclr} is also to tweak the network so that the softmax becomes more informative for OOD detection. Similarly to ReAct, one only has to get each class's softmax to make the score class-dependent.

\subsection{KNN}

For each $\{\vx_1,..., \vx_n\}$ from the training set, consider $\mH = \{h(\vx_1),...,h(\vx_n)\}$ where $h:\sX \rightarrow \R^p$ is defined such that $h(\vx_i)$ is the activation vector of $\vx_i$ of $f$'s penultimate layer. Let $N_{\mH}: \R^p \rightarrow \R^p$ be the nearest neighbor map such that $N(\vh)$ is the nearest neighbor of $\vh$ among $\mH$. KNN \cite{sun2022knnood} builds the score $\sood$ as
$$
\sood(\vx_{n+1}) = \| h(\vx_{n+1}) - N_{\mH}(h(\vx_{n+1}) \|.
$$

To make this score class-dependent, one can build $C$ maps $\{N_{\mH_1},...,N_{\mH_C}\}$ where $\mH_k = \{h(\vx_i) | f(\vx_i) = k\}$ and then define a new score
$$
\sood(\vx_{n+1}, y) = \| h(\vx_{n+1}) - N_{\mH_y}(h(\vx_{n+1}) \|
$$
\subsection{Mahalanobis}

Let consider the map $h$ as in KNN. For each $k \in \{1,...,C\}$, Mahalanobis distance method \cite{mahananobis18nips} computes $\Sigma_k$ and $\mu_k$, which are the empirical covariance matrix and mean vectors of each set of points $\{h(\vx_i)\}_{i | f(\vx_i) = k}$. Then, the score $\sood$ is computed as:

$$
\sood(\vx_{n+1}) =\sqrt{(\vx_{n+1}-\mu_{f(\vx_{n+1})})^{T}\Sigma^{-1}(\vx_{n+1}-\mu_{f(\vx_{n+1})})},
$$
where $\Sigma = \frac{1}{C}\sum_{k\in \{1,...,C\}}  \Sigma_k$. To make the score class-dependent, one simply has to define
$$
\sood(\vx_{n+1}, y) =\sqrt{(\vx_{n+1}-\mu_y)^{T}\Sigma_y^{-1}(\vx_{n+1}-\mu_y)}.
$$

\subsection{Gram}

Let $f$ be a classifier of depth $L$. Gram method \cite{gram20icml} builds a statistic $\delta: \sX \rightarrow \R^L$ that outputs the channel-wise correlation of the activation maps for each layer. First, $\{\delta(\vx_1),..., \delta(\vx_n)\}$ are computed. Then, a multi-dimensional statistic $\{d_{l,k}\}_{l\in \{1,...,L\}, k \in \{1,...,C\}}$ is computed for each layer after a class-wise aggregation. 

For a new test point $\vx_{n+1}$, $\delta(\vx_{n+1})$ is computed, along with $f(\vx_{n+1})$. The score is built out of a weighted mean of the layer-wise deviation:
$$
\sood(\vx_{n+1}) = \sum_{l \in \{1,...,L\}} w_l |\delta(\vx_{n+1})_l - d_{l, f(\vx_{n+1})}|,
$$
where $\{w_l\}_{l \in \{1,...,L\}}$ are some normalization weights computed with the training data. It is quite straightforward to make this OOD score class-dependent by defining

$$
\sood(\vx_{n+1}, y) = \sum_{l \in \{1,...,L\}} w_l |\delta(\vx_{n+1})_l - d_{l, y}|.
$$

\section{Appendix: Complementary results on OpenOOD benchmark}
\label{app:appB}

In this section, we present the full results of benchmarks on OpenOOD. The results displayed are AUROC with $\delta=0.05$ in Table \ref{tab:openood005auroc}, FPR@TPR95 with $\delta=0.05$ in Table \ref{tab:openood005fpr} and FPR@TPR95 with $\delta=0.01$ in Table \ref{tab:openood001fpr}.

  \begin{table*}[!h]
    \centering
    \resizebox{\textwidth}{!}{
    \begin{tabular}{l|cccc|cccc|cccc} 
        \toprule
         & \multicolumn{4}{c|}{\textbf{CIFAR-10}} & \multicolumn{4}{c|}{\textbf{CIFAR-100}} & \multicolumn{4}{c}{\textbf{ImageNet-200}} \\
        
        \midrule
        OOD type  & \multicolumn{2}{c}{Near} & \multicolumn{2}{c|}{Far}  & \multicolumn{2}{c}{Near} & \multicolumn{2}{c|}{Far}  & \multicolumn{2}{c}{Near} & \multicolumn{2}{c}{Far}  \\
           & marg. & conf. & marg. & conf. & marg. & conf. & marg. & conf. & marg. & conf. & marg. & conf. \\
        \midrule
OpenMax \cite{openmax16cvpr} & {87.2} & \textbf{86.18} & {89.53} & \textbf{88.52} & {76.66} & \textbf{75.26} & {79.12} & \textbf{77.81} & {80.4} & \textbf{79.2} & {90.41} & \textbf{89.49} \\ 

MSP \cite{msp17iclr} & {87.68} & \textbf{86.76} & {91.0} & \textbf{90.17} & {80.42} & \textbf{79.2} & {77.58} & \textbf{76.29} & {83.3} & \textbf{82.2} & {90.2} & \textbf{89.36} \\ 

TempScale \cite{guo2017calibration} & {87.65} & \textbf{86.75} & {91.27} & \textbf{90.48} & {80.98} & \textbf{79.78} & {78.51} & \textbf{77.24} & {83.66} & \textbf{82.58} & {90.91} & \textbf{90.11} \\ 

ODIN \cite{odin18iclr} & {80.25} & \textbf{79.26} & {87.21} & \textbf{86.43} & {79.8} & \textbf{78.57} & {79.44} & \textbf{78.2} & {80.32} & \textbf{79.19} & {91.89} & \textbf{91.17} \\ 

MDS \cite{mahananobis18nips} & {86.72} & \textbf{85.71} & {90.2} & \textbf{89.29} & {58.79} & \textbf{57.2} & {70.06} & \textbf{68.63} & {62.51} & \textbf{60.96} & {74.94} & \textbf{73.6} \\ 

MDSEns \cite{mahananobis18nips} & {60.46} & \textbf{59.01} & {74.07} & \textbf{72.96} & {45.98} & \textbf{44.34} & {66.03} & \textbf{64.72} & {54.58} & \textbf{52.99} & {70.08} & \textbf{68.76} \\ 

Gram \cite{gram20icml} & {52.63} & \textbf{51.04} & {69.74} & \textbf{68.41} & {50.69} & \textbf{49.06} & {73.97} & \textbf{72.87} & {68.36} & \textbf{67.0} & {70.94} & \textbf{69.69} \\ 

EBO \cite{energyood20nips} & {86.93} & \textbf{86.08} & {91.74} & \textbf{91.05} & {80.84} & \textbf{79.63} & {79.71} & \textbf{78.47} & {82.57} & \textbf{81.47} & {91.12} & \textbf{90.33} \\ 

GradNorm \cite{huang2021importance} & {53.77} & \textbf{52.26} & {58.55} & \textbf{57.09} & {69.73} & \textbf{68.41} & {68.82} & \textbf{67.48} & {73.33} & \textbf{72.12} & {85.29} & \textbf{84.45} \\ 

ReAct \cite{react21nips} & {86.47} & \textbf{85.6} & {91.02} & \textbf{90.28} & {80.7} & \textbf{79.5} & {79.84} & \textbf{78.6} & {80.48} & \textbf{79.35} & {93.1} & \textbf{92.4} \\ 

MLS \cite{species22icml} & {86.86} & \textbf{86.0} & {91.61} & \textbf{90.9} & {81.04} & \textbf{79.84} & {79.6} & \textbf{78.35} & {82.96} & \textbf{81.88} & {91.34} & \textbf{90.56} \\ 

KLM \cite{species22icml}  & {78.8} & \textbf{77.8} & {82.76} & \textbf{81.83} & {76.9} & \textbf{75.65} & {76.03} & \textbf{74.8} & {80.69} & \textbf{79.54} & {88.41} & \textbf{87.44} \\ 

VIM \cite{haoqi2022vim}  & {88.51} & \textbf{87.62} & {93.14} & \textbf{92.41} & {74.83} & \textbf{73.47} & {82.11} & \textbf{80.95} & {78.81} & \textbf{77.57} & {91.52} & \textbf{90.7} \\ 

KNN \cite{sun2022knnood} & {90.7} & \textbf{89.87} & {93.1} & \textbf{92.35} & {80.25} & \textbf{79.05} & {82.32} & \textbf{81.19} & {81.75} & \textbf{80.63} & {93.47} & \textbf{92.83} \\ 

DICE \cite{sun2021dice} & {77.79} & \textbf{76.68} & {85.41} & \textbf{84.56} & {79.15} & \textbf{77.89} & {79.84} & \textbf{78.61} & {81.97} & \textbf{80.86} & {91.19} & \textbf{90.43} \\ 

RankFeat \cite{song2022rankfeat} & {76.33} & \textbf{75.05} & {70.15} & \textbf{68.71} & {62.22} & \textbf{60.67} & {67.74} & \textbf{66.24} & {58.57} & \textbf{57.06} & {38.97} & \textbf{37.43} \\ 

ASH \cite{djurisic2023extremely} & {74.11} & \textbf{72.96} & {78.36} & \textbf{77.27} & {78.39} & \textbf{77.16} & {79.7} & \textbf{78.5} & {82.12} & \textbf{81.07} & {94.23} & \textbf{93.66} \\ 

SHE \cite{she23iclr} & {80.84} & \textbf{79.86} & {86.55} & \textbf{85.73} & {78.72} & \textbf{77.46} & {77.35} & \textbf{76.08} & {80.46} & \textbf{79.34} & {90.48} & \textbf{89.72} \\     
        \bottomrule
        \end{tabular}
    }
    \caption{Classical AUROC (marg.) vs Conformal AUROC (conf.) obtained with the Monte Carlo method and $\delta=0.05$ for several baselines from OpenOOD benchmark.}
    \label{tab:openood005auroc}
    \end{table*}

  \begin{table*}[h]
    \centering
    \resizebox{\textwidth}{!}{
    \begin{tabular}{l|cccc|cccc|cccc} 
        \toprule
         & \multicolumn{4}{c|}{\textbf{CIFAR-10}} & \multicolumn{4}{c|}{\textbf{CIFAR-100}} & \multicolumn{4}{c}{\textbf{ImageNet-200}} \\
        
        \midrule
        OOD type  & \multicolumn{2}{c}{Near} & \multicolumn{2}{c|}{Far}  & \multicolumn{2}{c}{Near} & \multicolumn{2}{c|}{Far}  & \multicolumn{2}{c}{Near} & \multicolumn{2}{c}{Far}  \\
           & marg. & conf. & marg. & conf. & marg. & conf. & marg. & conf. & marg. & conf. & marg. & conf. \\
        \midrule
OpenMax \cite{openmax16cvpr} & {46.77} & \textbf{48.98} & {29.48} & \textbf{31.48} & {55.57} & \textbf{57.8} & {54.77} & \textbf{57.0} & {63.32} & \textbf{65.75} & {32.29} & \textbf{35.35} \\ 

MSP \cite{msp17iclr} & {53.57} & \textbf{55.8} & {31.44} & \textbf{33.45} & {54.73} & \textbf{56.96} & {59.08} & \textbf{61.31} & {55.25} & \textbf{57.69} & {35.44} & \textbf{38.29} \\ 

TempScale \cite{guo2017calibration} & {56.85} & \textbf{59.08} & {33.36} & \textbf{35.38} & {54.77} & \textbf{56.99} & {58.24} & \textbf{60.47} & {55.03} & \textbf{57.5} & {34.11} & \textbf{37.06} \\ 

ODIN \cite{odin18iclr} & {84.55} & \textbf{86.78} & {60.9} & \textbf{62.97} & {58.44} & \textbf{60.67} & {57.75} & \textbf{59.98} & {66.38} & \textbf{68.8} & {33.66} & \textbf{36.75} \\ 

MDS \cite{mahananobis18nips} & {46.22} & \textbf{48.44} & {30.3} & \textbf{32.3} & {82.75} & \textbf{84.98} & {70.46} & \textbf{72.68} & {79.34} & \textbf{81.52} & {61.26} & \textbf{63.81} \\ 

MDSEns \cite{mahananobis18nips} & {92.06} & \textbf{94.29} & {61.09} & \textbf{62.87} & {95.84} & \textbf{98.07} & {66.97} & \textbf{68.85} & {91.69} & \textbf{93.8} & {80.43} & \textbf{82.89} \\ 

Gram \cite{gram20icml} & {93.52} & \textbf{95.75} & {69.29} & \textbf{71.48} & {92.48} & \textbf{94.71} & {63.1} & \textbf{65.2} & {85.43} & \textbf{87.63} & {84.95} & \textbf{87.44} \\ 

EBO \cite{energyood20nips} & {67.54} & \textbf{69.77} & {40.55} & \textbf{42.58} & {55.49} & \textbf{57.72} & {56.41} & \textbf{58.64} & {59.46} & \textbf{61.93} & {34.0} & \textbf{37.07} \\ 

GradNorm \cite{huang2021importance} & {95.37} & \textbf{97.6} & {89.34} & \textbf{91.52} & {86.13} & \textbf{88.36} & {82.79} & \textbf{85.02} & {83.07} & \textbf{85.33} & {66.78} & \textbf{69.67} \\ 

ReAct \cite{react21nips} & {71.56} & \textbf{73.78} & {42.43} & \textbf{44.52} & {56.74} & \textbf{58.97} & {56.32} & \textbf{58.55} & {65.37} & \textbf{67.8} & {27.21} & \textbf{30.28} \\ 

MLS \cite{species22icml} & {67.54} & \textbf{69.77} & {40.53} & \textbf{42.56} & {55.48} & \textbf{57.71} & {56.53} & \textbf{58.76} & {58.94} & \textbf{61.44} & {33.59} & \textbf{36.68} \\ 

KLM \cite{species22icml}  & {86.41} & \textbf{88.63} & {76.42} & \textbf{78.65} & {79.52} & \textbf{81.75} & {70.16} & \textbf{72.39} & {69.42} & \textbf{71.91} & {39.57} & \textbf{42.56} \\ 

VIM \cite{haoqi2022vim}  & {48.07} & \textbf{50.29} & {25.77} & \textbf{27.65} & {62.96} & \textbf{65.19} & {49.72} & \textbf{51.95} & {59.91} & \textbf{62.32} & {26.86} & \textbf{29.81} \\ 

KNN \cite{sun2022knnood} & {34.54} & \textbf{36.65} & {23.88} & \textbf{25.77} & {61.32} & \textbf{63.54} & {54.04} & \textbf{56.27} & {60.42} & \textbf{62.9} & {26.49} & \textbf{29.66} \\ 

DICE \cite{sun2021dice} & {80.15} & \textbf{82.38} & {53.93} & \textbf{56.06} & {58.1} & \textbf{60.33} & {55.95} & \textbf{58.17} & {60.98} & \textbf{63.46} & {35.93} & \textbf{39.04} \\ 

RankFeat \cite{song2022rankfeat} & {67.38} & \textbf{69.61} & {68.24} & \textbf{70.47} & {79.94} & \textbf{82.17} & {68.89} & \textbf{71.11} & {92.02} & \textbf{93.91} & {98.48} & \textbf{99.58} \\ 

ASH \cite{djurisic2023extremely} & {89.03} & \textbf{91.26} & {76.66} & \textbf{78.89} & {66.14} & \textbf{68.37} & {62.67} & \textbf{64.89} & {65.95} & \textbf{68.44} & {26.26} & \textbf{29.46} \\ 

SHE \cite{she23iclr} & {84.49} & \textbf{86.72} & {63.26} & \textbf{65.41} & {59.32} & \textbf{61.54} & {62.74} & \textbf{64.97} & {65.92} & \textbf{68.31} & {41.5} & \textbf{44.62} \\ 
        \bottomrule
        \end{tabular}
    }
    \caption{Classical FPR@TPR95 (marg.) vs Conformal FPR@TPR95 (conf.) obtained with the Monte Carlo method and $\delta=0.01$ for several baselines from OpenOOD benchmark.}
    \label{tab:openood001fpr}
    \end{table*}

  \begin{table*}[h]
    \centering
    \resizebox{\textwidth}{!}{
    \begin{tabular}{l|cccc|cccc|cccc} 
        \toprule
         & \multicolumn{4}{c|}{\textbf{CIFAR-10}} & \multicolumn{4}{c|}{\textbf{CIFAR-100}} & \multicolumn{4}{c}{\textbf{ImageNet-200}} \\
        
        \midrule
        OOD type  & \multicolumn{2}{c}{Near} & \multicolumn{2}{c|}{Far}  & \multicolumn{2}{c}{Near} & \multicolumn{2}{c|}{Far}  & \multicolumn{2}{c}{Near} & \multicolumn{2}{c}{Far}  \\
           & marg. & conf. & marg. & conf. & marg. & conf. & marg. & conf. & marg. & conf. & marg. & conf. \\
        \midrule
OpenMax \cite{openmax16cvpr} & {46.77} & \textbf{48.58} & {29.48} & \textbf{31.11} & {55.57} & \textbf{57.39} & {54.77} & \textbf{56.59} & {63.32} & \textbf{65.14} & {32.29} & \textbf{33.98} \\ 

MSP \cite{msp17iclr} & {53.57} & \textbf{55.39} & {31.44} & \textbf{33.08} & {54.73} & \textbf{56.55} & {59.08} & \textbf{60.9} & {55.25} & \textbf{57.06} & {35.44} & \textbf{37.16} \\ 

TempScale \cite{guo2017calibration} & {56.85} & \textbf{58.67} & {33.36} & \textbf{35.01} & {54.77} & \textbf{56.59} & {58.24} & \textbf{60.06} & {55.03} & \textbf{56.85} & {34.11} & \textbf{35.8} \\ 

ODIN \cite{odin18iclr} & {84.55} & \textbf{86.37} & {60.9} & \textbf{62.59} & {58.44} & \textbf{60.26} & {57.75} & \textbf{59.57} & {66.38} & \textbf{68.2} & {33.66} & \textbf{35.34} \\ 

MDS \cite{mahananobis18nips} & {46.22} & \textbf{48.03} & {30.3} & \textbf{31.94} & {82.75} & \textbf{84.57} & {70.46} & \textbf{72.28} & {79.34} & \textbf{81.16} & {61.26} & \textbf{63.08} \\ 

MDSEns \cite{mahananobis18nips} & {92.06} & \textbf{93.88} & {61.09} & \textbf{62.54} & {95.84} & \textbf{97.66} & {66.97} & \textbf{68.5} & {91.69} & \textbf{93.51} & {80.43} & \textbf{82.25} \\ 

Gram \cite{gram20icml} & {93.52} & \textbf{95.34} & {69.29} & \textbf{71.08} & {92.48} & \textbf{94.3} & {63.1} & \textbf{64.81} & {85.43} & \textbf{87.25} & {84.95} & \textbf{86.77} \\ 

EBO \cite{energyood20nips} & {67.54} & \textbf{69.36} & {40.55} & \textbf{42.21} & {55.49} & \textbf{57.31} & {56.41} & \textbf{58.23} & {59.46} & \textbf{61.28} & {34.0} & \textbf{35.7} \\ 

GradNorm \cite{huang2021importance} & {95.37} & \textbf{97.19} & {89.34} & \textbf{91.16} & {86.13} & \textbf{87.95} & {82.79} & \textbf{84.61} & {83.07} & \textbf{84.89} & {66.78} & \textbf{68.6} \\ 

ReAct \cite{react21nips} & {71.56} & \textbf{73.38} & {42.43} & \textbf{44.14} & {56.74} & \textbf{58.56} & {56.32} & \textbf{58.14} & {65.37} & \textbf{67.19} & {27.21} & \textbf{28.81} \\ 

MLS \cite{species22icml} & {67.54} & \textbf{69.36} & {40.53} & \textbf{42.19} & {55.48} & \textbf{57.3} & {56.53} & \textbf{58.35} & {58.94} & \textbf{60.76} & {33.59} & \textbf{35.28} \\ 

KLM \cite{species22icml}  & {86.41} & \textbf{88.23} & {76.42} & \textbf{78.24} & {79.52} & \textbf{81.34} & {70.16} & \textbf{71.98} & {69.42} & \textbf{71.24} & {39.57} & \textbf{41.3} \\ 

VIM \cite{haoqi2022vim}  & {48.07} & \textbf{49.88} & {25.77} & \textbf{27.3} & {62.96} & \textbf{64.78} & {49.72} & \textbf{51.54} & {59.91} & \textbf{61.72} & {26.86} & \textbf{28.46} \\ 

KNN \cite{sun2022knnood} & {34.54} & \textbf{36.27} & {23.88} & \textbf{25.42} & {61.32} & \textbf{63.14} & {54.04} & \textbf{55.86} & {60.42} & \textbf{62.23} & {26.49} & \textbf{28.09} \\ 

DICE \cite{sun2021dice} & {80.15} & \textbf{81.97} & {53.93} & \textbf{55.67} & {58.1} & \textbf{59.92} & {55.95} & \textbf{57.77} & {60.98} & \textbf{62.8} & {35.93} & \textbf{37.66} \\ 

RankFeat \cite{song2022rankfeat} & {67.38} & \textbf{69.2} & {68.24} & \textbf{70.06} & {79.94} & \textbf{81.76} & {68.89} & \textbf{70.71} & {92.02} & \textbf{93.84} & {98.48} & \textbf{99.55} \\ 

ASH \cite{djurisic2023extremely} & {89.03} & \textbf{90.85} & {76.66} & \textbf{78.48} & {66.14} & \textbf{67.96} & {62.67} & \textbf{64.49} & {65.95} & \textbf{67.77} & {26.26} & \textbf{27.85} \\ 

SHE \cite{she23iclr} & {84.49} & \textbf{86.31} & {63.26} & \textbf{65.02} & {59.32} & \textbf{61.14} & {62.74} & \textbf{64.56} & {65.92} & \textbf{67.74} & {41.5} & \textbf{43.27} \\

        \bottomrule
        \end{tabular}
    }
    \caption{Classical FPR@TPR95 (marg.) vs Conformal FPR@TPR95 (conf.) obtained with the Monte Carlo method and $\delta=0.05$ for several baselines from OpenOOD benchmark.}
    \label{tab:openood005fpr}
    \end{table*}

\newpage
\section{Appendix: Full results for ADBench}
\label{app:appC}

In this section, we present the full results of the ADBench benchmark. Table \ref{tab:adbench_classical} displays classical AUROC, Table \ref{tab:adbench_conformal} displays conformal AUROC, and Table \ref{tab:adbench_correction} displays the difference between the two (AUROC correction), all with $\delta=0.05$.

\begin{table*}[h]
    \centering
    \resizebox{\textwidth}{!}{
    \begin{tabular}{lccccccccccccc} 
        \toprule
        & IForest & OCSVM & CBLOF & COF & COPOD & ECOD & HBOS & KNN &
       LOF & PCA & SOD & DeepSVDD & DAGMM\\
       \midrule
       cover & 0.87 & 0.93 & 0.89 & 0.77 & 0.89 & 0.92 & 0.80 & 0.86 & 0.85 & 0.94 & 0.74 & 0.46 & 0.90\\
       donors & 0.78 & 0.72 & 0.62 & 0.71 & 0.82 & 0.89 & 0.78 & 0.82 & 0.59 & 0.83 & 0.56 & 0.36 & 0.71\\
       fault & 0.57 & 0.48 & 0.64 & 0.62 & 0.44 & 0.45 & 0.51 & 0.73 & 0.59 & 0.46 & 0.68 & 0.52 & 0.46\\
       fraud & 0.90 & 0.91 & 0.88 & 0.96 & 0.88 & 0.89 & 0.90 & 0.93 & 0.96 & 0.90 & 0.95 & 0.73 & 0.90\\
       glass & 0.77 & 0.35 & 0.83 & 0.72 & 0.72 & 0.66 & 0.77 & 0.82 & 0.69 & 0.66 & 0.73 & 0.47 & 0.76\\
       Hepatitis & 0.70 & 0.68 & 0.66 & 0.41 & 0.82 & 0.75 & 0.80 & 0.53 & 0.38 & 0.76 & 0.68 & 0.52 & 0.55\\
       Ionosphere & 0.84 & 0.76 & 0.91 & 0.87 & 0.79 & 0.73 & 0.62 & 0.88 & 0.91 & 0.79 & 0.86 & 0.51 & 0.73\\
       landsat & 0.48 & 0.36 & 0.64 & 0.53 & 0.42 & 0.36 & 0.55 & 0.58 & 0.54 & 0.36 & 0.60 & 0.63 & 0.44\\
       ALOI & 0.57 & 0.56 & 0.55 & 0.65 & 0.54 & 0.56 & 0.53 & 0.61 & 0.67 & 0.57 & 0.61 & 0.51 & 0.52\\
       letter & 0.61 & 0.46 & 0.76 & 0.80 & 0.54 & 0.56 & 0.60 & 0.86 & 0.84 & 0.50 & 0.84 & 0.56 & 0.50\\
       20news 0 & 0.64 & 0.63 & 0.71 & 0.71 & 0.61 & 0.61 & 0.62 & 0.73 & 0.80 & 0.64 & 0.73 & 0.50 & 0.63\\
       20news 1 & 0.51 & 0.53 & 0.52 & 0.58 & 0.52 & 0.54 & 0.53 & 0.57 & 0.61 & 0.54 & 0.58 & 0.48 & 0.54\\
       20news 2 & 0.50 & 0.51 & 0.47 & 0.53 & 0.50 & 0.52 & 0.51 & 0.51 & 0.54 & 0.51 & 0.50 & 0.49 & 0.53\\
       20news 3 & 0.75 & 0.72 & 0.83 & 0.81 & 0.75 & 0.75 & 0.74 & 0.79 & 0.71 & 0.73 & 0.70 & 0.67 & 0.54\\
       20news 4 & 0.48 & 0.51 & 0.45 & 0.57 & 0.48 & 0.51 & 0.50 & 0.48 & 0.51 & 0.51 & 0.53 & 0.53 & 0.48\\
       20news 5 & 0.52 & 0.49 & 0.47 & 0.50 & 0.48 & 0.46 & 0.49 & 0.48 & 0.55 & 0.48 & 0.48 & 0.49 & 0.54\\
       Lymphography & 1.00 & 1.00 & 1.00 & 0.91 & 0.99 & 1.00 & 0.99 & 0.56 & 0.90 & 1.00 & 0.73 & 0.34 & 0.72\\
       magic.gamma & 0.73 & 0.61 & 0.75 & 0.67 & 0.68 & 0.64 & 0.71 & 0.82 & 0.69 & 0.67 & 0.75 & 0.60 & 0.59\\
       musk & 1.00 & 0.81 & 1.00 & 0.39 & 0.94 & 0.95 & 1.00 & 0.70 & 0.41 & 1.00 & 0.74 & 0.56 & 0.77\\
       PageBlocks & 0.90 & 0.89 & 0.85 & 0.73 & 0.88 & 0.92 & 0.81 & 0.82 & 0.76 & 0.91 & 0.78 & 0.59 & 0.90\\
       pendigits & 0.95 & 0.94 & 0.90 & 0.45 & 0.91 & 0.93 & 0.93 & 0.73 & 0.48 & 0.94 & 0.66 & 0.42 & 0.64\\
       Pima & 0.73 & 0.67 & 0.71 & 0.61 & 0.69 & 0.63 & 0.71 & 0.73 & 0.66 & 0.71 & 0.61 & 0.51 & 0.56\\
       annthyroid & 0.82 & 0.57 & 0.62 & 0.66 & 0.77 & 0.79 & 0.60 & 0.72 & 0.70 & 0.66 & 0.77 & 0.77 & 0.57\\
       satellite & 0.70 & 0.59 & 0.71 & 0.55 & 0.63 & 0.58 & 0.75 & 0.65 & 0.56 & 0.60 & 0.64 & 0.55 & 0.62\\
       satimage-2 & 0.99 & 0.97 & 1.00 & 0.57 & 0.97 & 0.96 & 0.98 & 0.93 & 0.47 & 0.98 & 0.83 & 0.49 & 0.96\\
       shuttle & 1.00 & 0.97 & 0.83 & 0.52 & 0.99 & 0.99 & 0.99 & 0.70 & 0.57 & 0.99 & 0.70 & 0.49 & 0.98\\
       smtp & 0.86 & 0.72 & 0.70 & 0.69 & 0.70 & 0.78 & 0.56 & 0.84 & 0.58 & 0.83 & 0.40 & 0.72 & 0.71\\
       speech & 0.51 & 0.50 & 0.51 & 0.56 & 0.53 & 0.51 & 0.51 & 0.51 & 0.52 & 0.51 & 0.56 & 0.54 & 0.53\\
       Stamps & 0.91 & 0.84 & 0.68 & 0.54 & 0.93 & 0.88 & 0.91 & 0.69 & 0.51 & 0.91 & 0.73 & 0.56 & 0.89\\
       thyroid & 0.98 & 0.88 & 0.95 & 0.91 & 0.94 & 0.98 & 0.96 & 0.96 & 0.87 & 0.96 & 0.93 & 0.49 & 0.80\\
       vertebral & 0.37 & 0.38 & 0.41 & 0.49 & 0.26 & 0.41 & 0.29 & 0.34 & 0.49 & 0.37 & 0.40 & 0.37 & 0.53\\
       vowels & 0.75 & 0.63 & 0.90 & 0.95 & 0.55 & 0.62 & 0.73 & 0.97 & 0.93 & 0.67 & 0.92 & 0.56 & 0.61\\
       Waveform & 0.71 & 0.56 & 0.72 & 0.73 & 0.75 & 0.62 & 0.69 & 0.74 & 0.73 & 0.65 & 0.69 & 0.56 & 0.49\\
       WDBC & 0.99 & 0.99 & 0.99 & 0.96 & 0.99 & 0.97 & 0.99 & 0.92 & 0.89 & 0.99 & 0.92 & 0.62 & 0.77\\
       Wilt & 0.42 & 0.31 & 0.33 & 0.50 & 0.33 & 0.36 & 0.32 & 0.48 & 0.51 & 0.20 & 0.53 & 0.46 & 0.37\\
       wine & 0.80 & 0.73 & 0.26 & 0.44 & 0.89 & 0.77 & 0.91 & 0.45 & 0.38 & 0.84 & 0.46 & 0.60 & 0.62\\
       WPBC & 0.47 & 0.45 & 0.45 & 0.46 & 0.49 & 0.47 & 0.51 & 0.47 & 0.41 & 0.46 & 0.51 & 0.50 & 0.48\\
       yeast & 0.38 & 0.41 & 0.45 & 0.44 & 0.37 & 0.44 & 0.40 & 0.39 & 0.45 & 0.41 & 0.42 & 0.48 & 0.41\\
       campaign & 0.73 & 0.67 & 0.64 & 0.58 & 0.78 & 0.77 & 0.79 & 0.73 & 0.59 & 0.73 & 0.69 & 0.53 & 0.58\\
       cardio & 0.93 & 0.94 & 0.90 & 0.71 & 0.92 & 0.94 & 0.85 & 0.77 & 0.66 & 0.96 & 0.73 & 0.58 & 0.75\\
       Cardiotocography & 0.68 & 0.78 & 0.65 & 0.54 & 0.67 & 0.78 & 0.61 & 0.56 & 0.60 & 0.75 & 0.52 & 0.53 & 0.62\\
       celeba & 0.70 & 0.71 & 0.74 & 0.39 & 0.76 & 0.76 & 0.76 & 0.60 & 0.39 & 0.79 & 0.48 & 0.54 & 0.45\\
       CIFAR10 0 & 0.73 & 0.68 & 0.70 & 0.70 & 0.69 & 0.70 & 0.70 & 0.74 & 0.74 & 0.70 & 0.71 & 0.56 & 0.53\\
       CIFAR10 1 & 0.55 & 0.59 & 0.61 & 0.63 & 0.46 & 0.51 & 0.44 & 0.60 & 0.72 & 0.60 & 0.62 & 0.50 & 0.58\\
       CIFAR10 2 & 0.56 & 0.58 & 0.58 & 0.61 & 0.56 & 0.57 & 0.54 & 0.60 & 0.65 & 0.58 & 0.59 & 0.58 & 0.51\\
       CIFAR10 3 & 0.55 & 0.58 & 0.59 & 0.56 & 0.51 & 0.53 & 0.50 & 0.56 & 0.60 & 0.56 & 0.56 & 0.60 & 0.56\\
       CIFAR10 5 & 0.50 & 0.58 & 0.58 & 0.57 & 0.47 & 0.52 & 0.47 & 0.54 & 0.60 & 0.57 & 0.54 & 0.46 & 0.59\\
       CIFAR10 6 & 0.64 & 0.65 & 0.68 & 0.69 & 0.65 & 0.66 & 0.65 & 0.72 & 0.72 & 0.68 & 0.69 & 0.57 & 0.50\\
       CIFAR10 7 & 0.54 & 0.59 & 0.56 & 0.57 & 0.52 & 0.55 & 0.50 & 0.54 & 0.60 & 0.57 & 0.56 & 0.62 & 0.61\\
       agnews 0 & 0.50 & 0.47 & 0.54 & 0.61 & 0.49 & 0.47 & 0.48 & 0.58 & 0.63 & 0.47 & 0.56 & 0.35 & 0.48\\
       agnews 1 & 0.58 & 0.54 & 0.58 & 0.71 & 0.51 & 0.54 & 0.55 & 0.62 & 0.74 & 0.55 & 0.61 & 0.37 & 0.56\\
       agnews 2 & 0.65 & 0.61 & 0.71 & 0.73 & 0.61 & 0.59 & 0.61 & 0.75 & 0.79 & 0.61 & 0.73 & 0.50 & 0.53\\
       agnews 3 & 0.54 & 0.55 & 0.57 & 0.70 & 0.51 & 0.53 & 0.51 & 0.62 & 0.70 & 0.55 & 0.61 & 0.50 & 0.51\\
       amazon & 0.56 & 0.54 & 0.58 & 0.58 & 0.57 & 0.54 & 0.56 & 0.59 & 0.56 & 0.54 & 0.58 & 0.45 & 0.51\\
       imdb & 0.50 & 0.45 & 0.50 & 0.49 & 0.50 & 0.45 & 0.48 & 0.48 & 0.49 & 0.46 & 0.50 & 0.52 & 0.42\\
       yelp & 0.61 & 0.59 & 0.64 & 0.68 & 0.60 & 0.57 & 0.59 & 0.68 & 0.66 & 0.59 & 0.66 & 0.50 & 0.55\\       
       \bottomrule
    \end{tabular}
}
\caption{Full results for ADBench: classical AUROC.}
\label{tab:adbench_classical}
\end{table*}

\begin{table*}[ht]
    \centering
    \resizebox{\textwidth}{!}{
    \begin{tabular}{lccccccccccccc} 
        \toprule
        & IForest & OCSVM & CBLOF & COF & COPOD & ECOD & HBOS & KNN &
       LOF & PCA & SOD & DeepSVDD & DAGMM\\
       \midrule
       cover & 0.75 & 0.83 & 0.79 & 0.64 & 0.78 & 0.82 & 0.74 & 0.74 & 0.73 & 0.84 & 0.60 & 0.30 & 0.79\\
donors & 0.72 & 0.66 & 0.54 & 0.64 & 0.77 & 0.85 & 0.72 & 0.76 & 0.53 & 0.78 & 0.50 & 0.31 & 0.65\\
fault & 0.48 & 0.40 & 0.55 & 0.53 & 0.35 & 0.37 & 0.43 & 0.65 & 0.50 & 0.37 & 0.59 & 0.43 & 0.37\\
fraud & 0.77 & 0.63 & 0.63 & 0.82 & 0.62 & 0.64 & 0.86 & 0.68 & 0.79 & 0.66 & 0.69 & 0.51 & 0.64\\
glass & 0.65 & 0.31 & 0.73 & 0.60 & 0.60 & 0.53 & 0.64 & 0.72 & 0.58 & 0.54 & 0.63 & 0.38 & 0.66\\
Hepatitis & 0.54 & 0.52 & 0.52 & 0.23 & 0.70 & 0.61 & 0.66 & 0.26 & 0.22 & 0.62 & 0.54 & 0.38 & 0.40\\
Ionosphere & 0.77 & 0.68 & 0.85 & 0.80 & 0.70 & 0.63 & 0.50 & 0.83 & 0.85 & 0.71 & 0.80 & 0.38 & 0.64\\
landsat & 0.42 & 0.30 & 0.58 & 0.48 & 0.35 & 0.30 & 0.49 & 0.52 & 0.48 & 0.30 & 0.54 & 0.58 & 0.39\\
ALOI & 0.49 & 0.46 & 0.45 & 0.55 & 0.43 & 0.46 & 0.46 & 0.52 & 0.58 & 0.47 & 0.52 & 0.41 & 0.42\\
letter & 0.44 & 0.30 & 0.61 & 0.66 & 0.36 & 0.39 & 0.42 & 0.74 & 0.72 & 0.33 & 0.71 & 0.40 & 0.35\\
20news 0 & 0.51 & 0.49 & 0.57 & 0.60 & 0.48 & 0.46 & 0.47 & 0.62 & 0.69 & 0.50 & 0.61 & 0.36 & 0.49\\
20news 1 & 0.36 & 0.39 & 0.37 & 0.44 & 0.37 & 0.39 & 0.37 & 0.44 & 0.47 & 0.39 & 0.44 & 0.33 & 0.39\\
20news 2 & 0.34 & 0.36 & 0.34 & 0.39 & 0.34 & 0.36 & 0.35 & 0.36 & 0.38 & 0.36 & 0.34 & 0.32 & 0.38\\
20news 3 & 0.65 & 0.61 & 0.73 & 0.71 & 0.63 & 0.64 & 0.64 & 0.69 & 0.58 & 0.62 & 0.59 & 0.54 & 0.45\\
20news 4 & 0.28 & 0.31 & 0.27 & 0.39 & 0.27 & 0.32 & 0.30 & 0.30 & 0.34 & 0.31 & 0.35 & 0.35 & 0.30\\
20news 5 & 0.34 & 0.32 & 0.29 & 0.30 & 0.30 & 0.31 & 0.32 & 0.29 & 0.35 & 0.32 & 0.28 & 0.31 & 0.35\\
Lymphography & 0.95 & 0.95 & 0.95 & 0.83 & 0.95 & 0.95 & 0.95 & 0.45 & 0.81 & 0.96 & 0.62 & 0.25 & 0.62\\
magic.gamma & 0.70 & 0.57 & 0.72 & 0.63 & 0.65 & 0.60 & 0.68 & 0.79 & 0.65 & 0.64 & 0.72 & 0.56 & 0.55\\
musk & 0.57 & 0.65 & 0.22 & 0.21 & 0.83 & 0.85 & 0.58 & 0.53 & 0.22 & 0.32 & 0.59 & 0.42 & 0.64\\
PageBlocks & 0.84 & 0.83 & 0.79 & 0.67 & 0.82 & 0.86 & 0.74 & 0.76 & 0.70 & 0.85 & 0.71 & 0.52 & 0.84\\
pendigits & 0.87 & 0.86 & 0.82 & 0.33 & 0.82 & 0.85 & 0.85 & 0.60 & 0.35 & 0.86 & 0.53 & 0.29 & 0.52\\
Pima & 0.63 & 0.57 & 0.62 & 0.51 & 0.59 & 0.54 & 0.62 & 0.64 & 0.55 & 0.61 & 0.52 & 0.40 & 0.45\\
annthyroid & 0.76 & 0.49 & 0.55 & 0.59 & 0.70 & 0.72 & 0.53 & 0.65 & 0.63 & 0.59 & 0.71 & 0.71 & 0.49\\
satellite & 0.67 & 0.55 & 0.67 & 0.50 & 0.59 & 0.54 & 0.71 & 0.61 & 0.51 & 0.56 & 0.59 & 0.50 & 0.58\\
satimage-2 & 0.51 & 0.71 & 0.45 & 0.41 & 0.74 & 0.80 & 0.77 & 0.79 & 0.34 & 0.70 & 0.68 & 0.31 & 0.84\\
shuttle & 0.94 & 0.87 & 0.74 & 0.46 & 0.89 & 0.94 & 0.94 & 0.65 & 0.52 & 0.85 & 0.63 & 0.42 & 0.91\\
smtp & 0.81 & 0.60 & 0.58 & 0.59 & 0.58 & 0.68 & 0.45 & 0.76 & 0.48 & 0.72 & 0.32 & 0.60 & 0.60\\
speech & 0.31 & 0.31 & 0.31 & 0.37 & 0.34 & 0.32 & 0.31 & 0.31 & 0.33 & 0.32 & 0.35 & 0.34 & 0.34\\
Stamps & 0.84 & 0.75 & 0.57 & 0.42 & 0.87 & 0.80 & 0.83 & 0.57 & 0.39 & 0.85 & 0.62 & 0.42 & 0.81\\
thyroid & 0.90 & 0.77 & 0.86 & 0.81 & 0.86 & 0.90 & 0.92 & 0.87 & 0.77 & 0.88 & 0.84 & 0.33 & 0.69\\
vertebral & 0.23 & 0.26 & 0.28 & 0.37 & 0.14 & 0.29 & 0.17 & 0.20 & 0.37 & 0.25 & 0.29 & 0.24 & 0.40\\
vowels & 0.57 & 0.45 & 0.73 & 0.76 & 0.35 & 0.45 & 0.54 & 0.77 & 0.74 & 0.49 & 0.75 & 0.35 & 0.43\\
Waveform & 0.56 & 0.42 & 0.59 & 0.58 & 0.60 & 0.47 & 0.53 & 0.59 & 0.59 & 0.50 & 0.54 & 0.39 & 0.34\\
WDBC & 0.95 & 0.95 & 0.95 & 0.91 & 0.95 & 0.92 & 0.96 & 0.86 & 0.81 & 0.95 & 0.85 & 0.50 & 0.67\\
Wilt & 0.29 & 0.20 & 0.21 & 0.37 & 0.20 & 0.24 & 0.19 & 0.35 & 0.38 & 0.12 & 0.42 & 0.34 & 0.26\\
wine & 0.70 & 0.63 & 0.12 & 0.31 & 0.80 & 0.67 & 0.84 & 0.33 & 0.26 & 0.75 & 0.32 & 0.48 & 0.48\\
WPBC & 0.33 & 0.32 & 0.32 & 0.33 & 0.36 & 0.33 & 0.38 & 0.32 & 0.29 & 0.33 & 0.39 & 0.38 & 0.35\\
yeast & 0.29 & 0.31 & 0.35 & 0.35 & 0.28 & 0.34 & 0.31 & 0.30 & 0.36 & 0.31 & 0.32 & 0.38 & 0.31\\
campaign & 0.68 & 0.62 & 0.59 & 0.52 & 0.74 & 0.72 & 0.75 & 0.68 & 0.53 & 0.68 & 0.64 & 0.48 & 0.52\\
cardio & 0.84 & 0.85 & 0.80 & 0.60 & 0.83 & 0.84 & 0.75 & 0.67 & 0.53 & 0.86 & 0.62 & 0.47 & 0.64\\
Cardiotocography & 0.59 & 0.70 & 0.57 & 0.45 & 0.58 & 0.70 & 0.52 & 0.48 & 0.51 & 0.66 & 0.43 & 0.45 & 0.53\\
celeba & 0.64 & 0.65 & 0.69 & 0.30 & 0.70 & 0.71 & 0.71 & 0.28 & 0.30 & 0.74 & 0.38 & 0.48 & 0.37\\
CIFAR10 0 & 0.63 & 0.59 & 0.60 & 0.60 & 0.59 & 0.60 & 0.60 & 0.65 & 0.64 & 0.61 & 0.61 & 0.46 & 0.42\\
CIFAR10 1 & 0.43 & 0.48 & 0.50 & 0.52 & 0.35 & 0.39 & 0.33 & 0.49 & 0.62 & 0.49 & 0.51 & 0.39 & 0.48\\
CIFAR10 2 & 0.45 & 0.48 & 0.47 & 0.50 & 0.44 & 0.45 & 0.43 & 0.49 & 0.55 & 0.47 & 0.48 & 0.48 & 0.40\\
CIFAR10 3 & 0.44 & 0.48 & 0.49 & 0.46 & 0.40 & 0.42 & 0.39 & 0.47 & 0.50 & 0.46 & 0.46 & 0.49 & 0.45\\
CIFAR10 5 & 0.38 & 0.48 & 0.47 & 0.46 & 0.35 & 0.40 & 0.34 & 0.42 & 0.49 & 0.46 & 0.42 & 0.34 & 0.49\\
CIFAR10 6 & 0.54 & 0.55 & 0.58 & 0.59 & 0.54 & 0.55 & 0.54 & 0.61 & 0.62 & 0.58 & 0.59 & 0.47 & 0.39\\
CIFAR10 7 & 0.43 & 0.48 & 0.45 & 0.46 & 0.41 & 0.44 & 0.39 & 0.44 & 0.50 & 0.46 & 0.46 & 0.51 & 0.50\\
agnews 0 & 0.41 & 0.39 & 0.45 & 0.53 & 0.40 & 0.38 & 0.39 & 0.49 & 0.56 & 0.39 & 0.48 & 0.27 & 0.40\\
agnews 1 & 0.50 & 0.46 & 0.50 & 0.64 & 0.42 & 0.45 & 0.46 & 0.54 & 0.67 & 0.47 & 0.53 & 0.28 & 0.48\\
agnews 2 & 0.57 & 0.53 & 0.63 & 0.66 & 0.52 & 0.51 & 0.52 & 0.68 & 0.73 & 0.53 & 0.66 & 0.42 & 0.45\\
agnews 3 & 0.45 & 0.46 & 0.49 & 0.63 & 0.43 & 0.44 & 0.43 & 0.54 & 0.64 & 0.46 & 0.53 & 0.42 & 0.42\\
amazon & 0.48 & 0.45 & 0.50 & 0.49 & 0.48 & 0.46 & 0.47 & 0.50 & 0.48 & 0.46 & 0.50 & 0.37 & 0.43\\
imdb & 0.41 & 0.36 & 0.41 & 0.40 & 0.42 & 0.36 & 0.40 & 0.39 & 0.40 & 0.37 & 0.41 & 0.44 & 0.34\\
yelp & 0.52 & 0.50 & 0.55 & 0.60 & 0.52 & 0.49 & 0.51 & 0.60 & 0.59 & 0.51 & 0.58 & 0.42 & 0.47\\
       \bottomrule
    \end{tabular}
}
\caption{Full results for ADBench: conformal AUROC.}
\label{tab:adbench_conformal}
\end{table*}
\begin{table*}[ht]
    \centering
    \resizebox{\textwidth}{!}{
    \begin{tabular}{lccccccccccccc} 
        \toprule
        & IForest & OCSVM & CBLOF & COF & COPOD & ECOD & HBOS & KNN &
       LOF & PCA & SOD & DeepSVDD & DAGMM\\
       \midrule
       cover & 0.12 & 0.10 & 0.11 & 0.13 & 0.11 & 0.10 & 0.07 & 0.12 & 0.12 & 0.10 & 0.14 & 0.15 & 0.11\\
donors & 0.06 & 0.07 & 0.08 & 0.07 & 0.05 & 0.04 & 0.06 & 0.06 & 0.06 & 0.05 & 0.06 & 0.05 & 0.06\\
fault & 0.09 & 0.08 & 0.09 & 0.09 & 0.09 & 0.09 & 0.09 & 0.08 & 0.09 & 0.09 & 0.09 & 0.08 & 0.09\\
fraud & 0.13 & 0.28 & 0.25 & 0.14 & 0.26 & 0.25 & 0.04 & 0.26 & 0.16 & 0.25 & 0.26 & 0.22 & 0.26\\
glass & 0.12 & 0.05 & 0.10 & 0.12 & 0.12 & 0.13 & 0.13 & 0.10 & 0.11 & 0.13 & 0.10 & 0.09 & 0.10\\
Hepatitis & 0.16 & 0.15 & 0.14 & 0.18 & 0.12 & 0.14 & 0.13 & 0.27 & 0.16 & 0.14 & 0.14 & 0.14 & 0.14\\
Ionosphere & 0.08 & 0.08 & 0.06 & 0.07 & 0.09 & 0.10 & 0.12 & 0.05 & 0.05 & 0.08 & 0.06 & 0.13 & 0.09\\
landsat & 0.06 & 0.06 & 0.05 & 0.05 & 0.06 & 0.06 & 0.06 & 0.06 & 0.05 & 0.06 & 0.05 & 0.05 & 0.05\\
ALOI & 0.07 & 0.10 & 0.10 & 0.09 & 0.10 & 0.10 & 0.06 & 0.09 & 0.08 & 0.10 & 0.09 & 0.10 & 0.10\\
letter & 0.17 & 0.16 & 0.15 & 0.14 & 0.19 & 0.17 & 0.18 & 0.13 & 0.13 & 0.17 & 0.13 & 0.15 & 0.15\\
20news 0 & 0.14 & 0.14 & 0.13 & 0.11 & 0.14 & 0.15 & 0.14 & 0.11 & 0.11 & 0.14 & 0.12 & 0.13 & 0.14\\
20news 1 & 0.15 & 0.15 & 0.15 & 0.14 & 0.15 & 0.15 & 0.16 & 0.13 & 0.14 & 0.15 & 0.14 & 0.14 & 0.15\\
20news 2 & 0.16 & 0.15 & 0.14 & 0.15 & 0.16 & 0.16 & 0.16 & 0.15 & 0.16 & 0.15 & 0.15 & 0.16 & 0.15\\
20news 3 & 0.10 & 0.11 & 0.10 & 0.10 & 0.11 & 0.11 & 0.10 & 0.10 & 0.13 & 0.11 & 0.12 & 0.13 & 0.09\\
20news 4 & 0.20 & 0.20 & 0.17 & 0.18 & 0.21 & 0.19 & 0.20 & 0.18 & 0.16 & 0.20 & 0.17 & 0.18 & 0.18\\
20news 5 & 0.17 & 0.17 & 0.18 & 0.20 & 0.18 & 0.15 & 0.17 & 0.19 & 0.20 & 0.16 & 0.20 & 0.18 & 0.19\\
Lymphography & 0.05 & 0.05 & 0.05 & 0.08 & 0.04 & 0.05 & 0.05 & 0.11 & 0.09 & 0.04 & 0.11 & 0.09 & 0.11\\
magic.gamma & 0.03 & 0.04 & 0.03 & 0.04 & 0.03 & 0.04 & 0.03 & 0.03 & 0.04 & 0.03 & 0.03 & 0.04 & 0.04\\
musk & 0.43 & 0.15 & 0.78 & 0.18 & 0.11 & 0.11 & 0.42 & 0.17 & 0.19 & 0.68 & 0.16 & 0.14 & 0.13\\
PageBlocks & 0.06 & 0.06 & 0.06 & 0.06 & 0.06 & 0.06 & 0.07 & 0.06 & 0.06 & 0.06 & 0.07 & 0.06 & 0.05\\
pendigits & 0.08 & 0.08 & 0.09 & 0.12 & 0.08 & 0.08 & 0.08 & 0.13 & 0.13 & 0.08 & 0.13 & 0.13 & 0.12\\
Pima & 0.10 & 0.10 & 0.10 & 0.10 & 0.10 & 0.09 & 0.09 & 0.10 & 0.10 & 0.09 & 0.10 & 0.11 & 0.11\\
annthyroid & 0.06 & 0.08 & 0.07 & 0.07 & 0.07 & 0.06 & 0.07 & 0.07 & 0.07 & 0.07 & 0.06 & 0.06 & 0.08\\
satellite & 0.04 & 0.04 & 0.04 & 0.05 & 0.04 & 0.04 & 0.04 & 0.05 & 0.05 & 0.04 & 0.05 & 0.05 & 0.04\\
satimage-2 & 0.48 & 0.26 & 0.55 & 0.16 & 0.23 & 0.16 & 0.21 & 0.14 & 0.13 & 0.27 & 0.15 & 0.18 & 0.12\\
shuttle & 0.06 & 0.10 & 0.09 & 0.06 & 0.10 & 0.05 & 0.05 & 0.04 & 0.05 & 0.13 & 0.06 & 0.07 & 0.07\\
smtp & 0.05 & 0.12 & 0.12 & 0.10 & 0.12 & 0.11 & 0.10 & 0.09 & 0.09 & 0.11 & 0.08 & 0.12 & 0.11\\
speech & 0.19 & 0.19 & 0.19 & 0.19 & 0.19 & 0.19 & 0.19 & 0.20 & 0.19 & 0.19 & 0.21 & 0.21 & 0.18\\
Stamps & 0.07 & 0.09 & 0.11 & 0.11 & 0.06 & 0.08 & 0.07 & 0.12 & 0.12 & 0.07 & 0.11 & 0.14 & 0.07\\
thyroid & 0.08 & 0.10 & 0.09 & 0.10 & 0.08 & 0.08 & 0.04 & 0.09 & 0.10 & 0.08 & 0.09 & 0.16 & 0.11\\
vertebral & 0.14 & 0.12 & 0.13 & 0.12 & 0.12 & 0.12 & 0.12 & 0.14 & 0.13 & 0.12 & 0.11 & 0.12 & 0.13\\
vowels & 0.19 & 0.18 & 0.17 & 0.20 & 0.20 & 0.17 & 0.20 & 0.20 & 0.19 & 0.18 & 0.17 & 0.21 & 0.19\\
Waveform & 0.15 & 0.15 & 0.14 & 0.15 & 0.15 & 0.15 & 0.15 & 0.14 & 0.14 & 0.16 & 0.15 & 0.16 & 0.16\\
WDBC & 0.04 & 0.04 & 0.04 & 0.06 & 0.04 & 0.05 & 0.04 & 0.06 & 0.08 & 0.04 & 0.07 & 0.12 & 0.10\\
Wilt & 0.13 & 0.11 & 0.12 & 0.12 & 0.13 & 0.12 & 0.14 & 0.13 & 0.13 & 0.08 & 0.12 & 0.12 & 0.11\\
wine & 0.10 & 0.10 & 0.14 & 0.14 & 0.09 & 0.11 & 0.08 & 0.12 & 0.11 & 0.10 & 0.15 & 0.11 & 0.13\\
WPBC & 0.13 & 0.13 & 0.13 & 0.13 & 0.13 & 0.13 & 0.14 & 0.14 & 0.12 & 0.13 & 0.12 & 0.12 & 0.13\\
yeast & 0.09 & 0.10 & 0.10 & 0.10 & 0.09 & 0.09 & 0.09 & 0.09 & 0.10 & 0.10 & 0.11 & 0.10 & 0.10\\
campaign & 0.05 & 0.05 & 0.05 & 0.06 & 0.05 & 0.05 & 0.04 & 0.05 & 0.06 & 0.05 & 0.06 & 0.05 & 0.06\\
cardio & 0.09 & 0.09 & 0.10 & 0.12 & 0.09 & 0.09 & 0.10 & 0.10 & 0.14 & 0.10 & 0.11 & 0.11 & 0.11\\
Cardiotocography & 0.09 & 0.08 & 0.08 & 0.09 & 0.09 & 0.08 & 0.08 & 0.08 & 0.09 & 0.09 & 0.08 & 0.08 & 0.09\\
celeba & 0.06 & 0.06 & 0.05 & 0.09 & 0.05 & 0.05 & 0.05 & 0.31 & 0.09 & 0.05 & 0.10 & 0.06 & 0.08\\
CIFAR10 0 & 0.10 & 0.10 & 0.10 & 0.10 & 0.10 & 0.10 & 0.10 & 0.10 & 0.10 & 0.10 & 0.10 & 0.10 & 0.11\\
CIFAR10 1 & 0.12 & 0.11 & 0.11 & 0.11 & 0.11 & 0.12 & 0.11 & 0.11 & 0.10 & 0.11 & 0.10 & 0.10 & 0.11\\
CIFAR10 2 & 0.11 & 0.11 & 0.11 & 0.11 & 0.11 & 0.11 & 0.11 & 0.11 & 0.11 & 0.11 & 0.11 & 0.10 & 0.11\\
CIFAR10 3 & 0.11 & 0.11 & 0.10 & 0.10 & 0.11 & 0.11 & 0.11 & 0.10 & 0.10 & 0.11 & 0.10 & 0.11 & 0.11\\
CIFAR10 5 & 0.12 & 0.10 & 0.11 & 0.11 & 0.12 & 0.12 & 0.12 & 0.12 & 0.11 & 0.11 & 0.12 & 0.12 & 0.10\\
CIFAR10 6 & 0.11 & 0.10 & 0.11 & 0.10 & 0.11 & 0.11 & 0.11 & 0.10 & 0.10 & 0.10 & 0.10 & 0.10 & 0.11\\
CIFAR10 7 & 0.11 & 0.11 & 0.10 & 0.10 & 0.11 & 0.11 & 0.11 & 0.10 & 0.10 & 0.10 & 0.10 & 0.11 & 0.11\\
agnews 0 & 0.09 & 0.09 & 0.08 & 0.08 & 0.09 & 0.09 & 0.09 & 0.08 & 0.08 & 0.09 & 0.08 & 0.09 & 0.09\\
agnews 1 & 0.09 & 0.09 & 0.08 & 0.07 & 0.09 & 0.09 & 0.09 & 0.08 & 0.07 & 0.09 & 0.08 & 0.09 & 0.09\\
agnews 2 & 0.08 & 0.08 & 0.08 & 0.07 & 0.08 & 0.09 & 0.08 & 0.07 & 0.06 & 0.08 & 0.07 & 0.08 & 0.08\\
agnews 3 & 0.09 & 0.09 & 0.08 & 0.07 & 0.09 & 0.09 & 0.09 & 0.08 & 0.07 & 0.09 & 0.08 & 0.08 & 0.08\\
amazon & 0.08 & 0.08 & 0.08 & 0.08 & 0.08 & 0.08 & 0.08 & 0.09 & 0.09 & 0.08 & 0.09 & 0.08 & 0.08\\
imdb & 0.09 & 0.09 & 0.09 & 0.09 & 0.09 & 0.09 & 0.09 & 0.09 & 0.09 & 0.09 & 0.09 & 0.08 & 0.08\\
yelp & 0.08 & 0.08 & 0.08 & 0.08 & 0.08 & 0.09 & 0.08 & 0.08 & 0.08 & 0.08 & 0.08 & 0.08 & 0.08\\

       \bottomrule
    \end{tabular}
}
\caption{Full results for ADBench: AUROC correction (difference between conformal and classical AUROC).}
\label{tab:adbench_correction}
\end{table*}

\end{document}